\pdfoutput=1

\documentclass[11pt]{article}

\usepackage{emnlp2021}

\usepackage{times}
\usepackage{latexsym}
\usepackage{graphicx}
\usepackage{url}
\usepackage{amsmath,amssymb}
\usepackage{multirow}
\usepackage{booktabs}
\usepackage{subcaption}
\usepackage{mdframed}
\usepackage{bbm}
\usepackage{arydshln}
\usepackage{tabularx}
\usepackage{caption}
\usepackage[most]{tcolorbox}
\usepackage{float}
\usepackage{fdsymbol}
\usepackage{lipsum}
\usepackage[T1]{fontenc}
\usepackage[utf8]{inputenc}
\usepackage{microtype}

\newcommand{\aggname}{\textsc{Coop}}
\newcommand{\scorename}{input-output word overlap}
\newcommand{\name}{\mbox{\sc BiMeanVAE}}
\newcommand{\simpleavg}{\mbox{SimpleAvg}}
\newcommand{\vaeorig}{TextVAE}
\newcommand{\yelp}{\mbox{\bf Yelp}}
\newcommand{\amazon}{\mbox{\bf Amazon}}

\newcommand{\hl}[1]{#1}
\newcommand{\hla}[1]{#1}

\usepackage[normalem]{ulem}
\newcommand{\gray}[1]{{\color[HTML]{808080} \sout{#1}}}

\definecolor{c1}{HTML}{0072B2}%
\definecolor{c2}{HTML}{D55E00}%
\definecolor{c3}{HTML}{009E73}%
\definecolor{c4}{HTML}{56B4E9}%
\definecolor{c5}{HTML}{CC79A7}%
\definecolor{c6}{HTML}{E69F00}%
\definecolor{c7}{HTML}{844E4D}%
\definecolor{c8}{HTML}{2D512A}%
\newcommand\one[1]{\textcolor{c1}{\textbf{#1}}}
\newcommand\two[1]{\textcolor{c2}{\textbf{#1}}}
\newcommand\three[1]{\textcolor{c3}{\textbf{#1}}}
\newcommand\four[1]{\textcolor{c4}{\textbf{#1}}}
\newcommand\five[1]{\textcolor{c5}{\textbf{#1}}}
\newcommand\six[1]{\textcolor{c6}{\textbf{#1}}}

\definecolor{darkgreen}{rgb}{0.0, 0.5, 0.0}

\title{Convex Aggregation for Opinion Summarization}

\author{
Hayate Iso$^{\spadesuit}$ \quad
Xiaolan Wang$^{\spadesuit}$ \\
\textbf{Yoshihiko Suhara$^{\spadesuit}$ \quad
Stefanos Angelidis$^{\heartsuit}$ \quad
Wang-Chiew Tan$^{\diamondsuit}$}\thanks{~~Work done while at Megagon Labs.} \\
$^\spadesuit$Megagon Labs \quad
$^\heartsuit$University of Edinburgh \quad
$^\diamondsuit$Facebook AI\\
\texttt{hayate@megagon.ai},~~\texttt{xiaolan@megagon.ai},\\
\texttt{yoshi@megagon.ai},~~\texttt{s.angelidis@ed.ac.uk},~~\texttt{wangchiew@fb.com}}

\begin{document}
\maketitle
\begin{abstract}
Recent advances in text autoencoders have significantly improved the quality of the latent space, which enables models to generate \hl{grammatical and consistent} text from aggregated latent vectors.
As a successful application of this property, unsupervised opinion summarization models generate a summary by decoding the aggregated latent vectors of inputs. More specifically, they perform the aggregation via {\em simple average.}
However, little is known about how the vector aggregation step affects the generation quality.
In this study, we revisit the commonly used {\em simple average} approach by examining the latent space and generated summaries. We found that text autoencoders tend to generate overly generic summaries from simply averaged latent vectors due to an unexpected $L_2$-norm shrinkage in the aggregated latent vectors, which we refer to as {\em summary vector degeneration}.
To overcome this issue, we develop a framework \aggname, which searches input combinations for the latent vector aggregation using {\em input-output word overlap}. Experimental results show that \aggname{} successfully alleviates the summary vector degeneration issue and establishes new state-of-the-art performance on two opinion summarization benchmarks.
Code is available at \url{https://github.com/megagonlabs/coop}.
\end{abstract}

\section{Introduction}
The unprecedented growth of online review platforms and the recent success of neural summarization techniques~\cite{cheng-lapata-2016-neural,see-etal-2017-get,liu-lapata-2019-hierarchical},
spurred significant interest in research on multi-document opinion summarization~\cite{angelidis-lapata-2018-summarizing,Chu:2019:MeanSum,brazinskas-etal-2020-unsupervised,suhara-etal-2020-opiniondigest,amplayo-lapata-2020-unsupervised,amplayo2021unsupervised}.
The goal of multi-document opinion summarization is
to generate a summary that represents salient opinions in the input reviews.

\begin{figure*}[t]
    \centering
    \includegraphics[width=\textwidth]{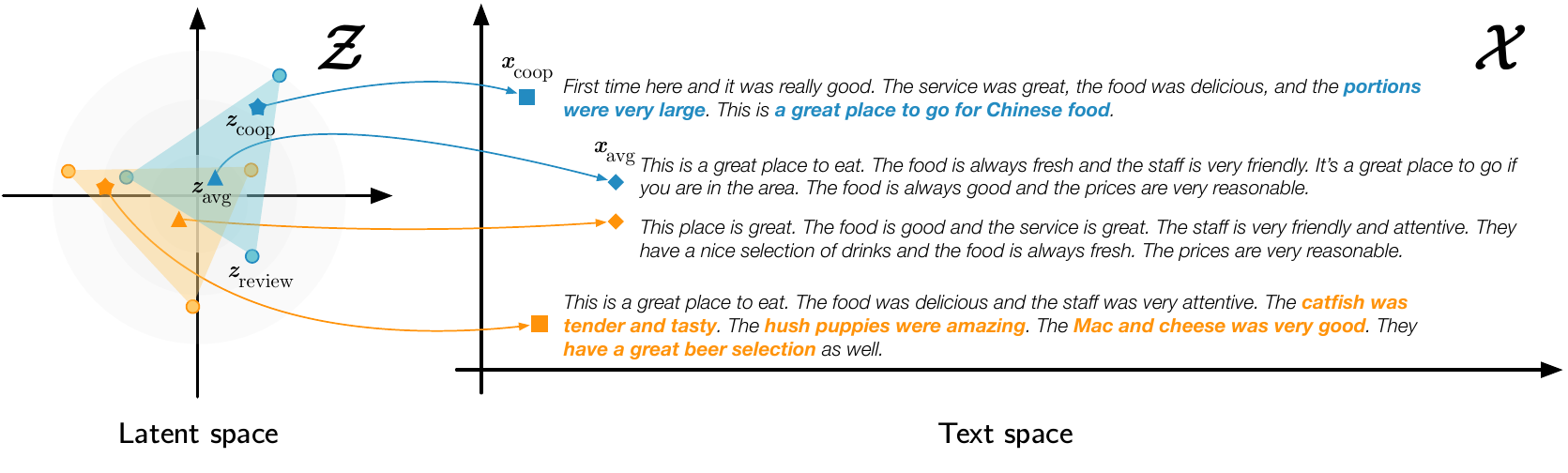}
    \caption{
    Illustration of the latent space $\mathcal{Z}$ and text space $\mathcal{X}$.
    The de facto standard approach in unsupervised opinion summarization uses the simple average of input review vectors $\boldsymbol{z}_{\text{review}}$ ($\circ$) to obtain the summary vector $\boldsymbol{z}_{\text{avg}}$ ($\smallblacktriangleup$).
    The simply averaged vector $\boldsymbol{z}_{\text{avg}}$ tends to be close to the center (i.e., has a small $L_2$-norm) in the latent space, and a generated summary $\boldsymbol{x}_{\text{avg}}$
    ($\smallblackdiamond$) tends to become overly generic.
    Our proposed framework \aggname{} finds a better aggregated vector to generate a more specific summary $\boldsymbol{x}_{\aggname{}}$($\smallblacksquare$) from the latent vector $\boldsymbol{z}_{\text{\aggname{}}}$ ($\star$).}
    \label{fig:overview}
\end{figure*}

Research on multi-document opinion summarization is challenging because of the lack of gold-standard summaries, which are difficult to collect at scale. This is in contrast to single-document summarization, where there exists an abundant annotated datasets~\cite{usnews,hermann2015teaching,rush-etal-2015-neural,narayan-etal-2018-dont}.
Consequently, the primary approach is to employ text autoencoders for unsupervised opinion summarization~\cite{Chu:2019:MeanSum,brazinskas-etal-2020-unsupervised}.
Text autoencoders, especially variational autoencoders (VAEs), are known for the ability to generate \hl{grammatical and consistent text by aggregating} multiple latent vectors~\cite{bowman-etal-2016-generating}.
Unsupervised opinion summarization models leverage this property to generate a summary by first aggregating the latent vectors of input reviews via {\em simple average}, and then decoding the summary from the aggregated vector.

However, it has not been verified if the simple average is the best choice for summary generation. Furthermore, little is known about the relationship between the latent vector and the generation quality. 
In this paper, we report that text autoencoder models with the simple average vector aggregation tend to generate overly generic summaries, which we refer to as {\em summary vector degeneration}. For example, as shown in Figure~\ref{fig:overview}, with simply averaged latent vectors, the generated summaries of two distinct entities are almost identical. 
We further discovered two factors that cause summary vector degeneration: (1) simply averaged latent vectors cause unexpected $L_2$-norm shrinkage, and (2) latent vectors with smaller $L_2$-norm are decoded into less informative summaries (e.g., contain only general information.)

To address the summary vector degeneration issue,
we develop \aggname{}, a latent vector aggregation framework.
In essence, \aggname{} considers {\it convex combinations} of the latent vectors of input reviews for better summary generation. More specifically, we focus on searching for a convex combination that maximizes the {\it input-output word overlap} between input reviews and a generated summary. This optimization strategy helps the model generate summaries that are more consistent with input reviews, thus improving the quality of summarization for unsupervised opinion summarization models.

Our contributions are summarized as follows:
\begin{itemize}
  \setlength{\parskip}{0cm}
  \setlength{\itemsep}{0cm}
  \item We report that the commonly used simple average vector aggregation method causes {\em summary vector degeneration}, which makes the decoder generate less informative and overly generic summaries.%
  \item We formalize \textit{latent vector aggregation} as an optimization problem, which considers the convex combination of input review latent vectors. We propose a solution, \aggname{}, to approximate the optimal latent vector with linear time complexity. To the best of our knowledge, this is the first work that optimizes latent vector aggregation for opinion summarization.
  \item We conduct comparative experiments against existing methods~\cite{Chu:2019:MeanSum, brazinskas-etal-2020-unsupervised}, which implement more sophisticated techniques. Our experiments demonstrate that by coupling with \aggname{}, two opinion summarization models (\name{} and Optimus) establish  new \mbox{state-of-the-art} performance on both \yelp\ and \amazon\ datasets.
\end{itemize}

\section{Preliminaries}\label{sec:preliminary}
Let us denote $\mathcal{R}=\{\boldsymbol{x}_i\}_{i=1}^{|\mathcal{R}|}$ as a dataset of customer reviews of the same domain (e.g., restaurant or product), where each
review is a sequence of words $\boldsymbol{x}=(x_1, ..., x_{\|\boldsymbol{x}\|})$ in the text space $\mathcal{X}$. Given an entity $e$ and its reviews $\mathcal{R}_e\subseteq \mathcal{R}$, the goal of the \textit{multi-document opinion summarization} task is to generate an abstractive summary $s_e$ such that the salient opinions in $\mathcal{R}_e$ are included.
\subsection{Unsupervised Opinion Summarization}~\label{sec:method}
Existing unsupervised opinion summarization models~\cite{Chu:2019:MeanSum, brazinskas-etal-2020-unsupervised} use the autoencoder, where an encoder $E: \mathcal{X} \xrightarrow{} \mathcal{Z}$ mapping from the text space $\mathcal{X}$ to latent space $\mathcal{Z}$, and a decoder $G: \mathcal{Z} \xrightarrow{} \mathcal{X}$ that generates texts from latent vectors.

\noindent
{\bf Encoder $E$}: Given an entity $e$ and its reviews $\mathcal{R}_e$, the encoder $E$ essentially maps every review $\boldsymbol{x}_i \in \mathcal{R}_e$ into the latent space: $\boldsymbol{z}_i = E(\boldsymbol{x}_i)$, where $\boldsymbol{z}_i$ is the latent vector of review $\boldsymbol{x}_i$.

\noindent
{\bf Decoder $G$}: The other core component is the decoder $G$, which generate a new text $\boldsymbol{\hat{x}} = (\hat{x}_1, ..., \hat{x}_{\|\boldsymbol{\hat{x}}\|})$ from a given latent vector $\boldsymbol{z}$: $\boldsymbol{\hat{x}} = G(\boldsymbol{z})$.

\noindent
{\bf Training}: At the training phase, the autoencoder model is trained to generate the review. While various training methods have been proposed, the simplest approach is aimed to reconstruct the input review from the corresponding latent vector.

\noindent
{\bf Generation}: At the generation phase, given a set of input review latent vectors $\mathcal{Z}_e = \{\boldsymbol{z}_1, ..., \boldsymbol{z}_{|\mathcal{R}_e|}\}$, existing opinion summarization models use simple average to create the latent vector of the summary ({\em summary vector}) 
$\boldsymbol{z}_{\texttt{summary}}^{\texttt{avg}} = \frac{1}{|\mathcal{R}_e|} \sum_{i = 1}^{|\mathcal{R}_e|} \boldsymbol{z}_i$, which is then decoded into the summary. In this paper, our focus is to analyze and improve the latent vector aggregation for the summary.

\subsection{Variational Autoencoders}\label{sec:vae}
In this study, we use variational autoencoders (VAEs) as the text autoencoder
since it provides a smooth latent space, which allows to \hl{produce grammatical and consistent text from aggregated latent vectors
~\cite{kingma2014auto,bowman-etal-2016-generating}}. More specifically, we tested two VAE variations, namely \name{} and Optimus~\cite{li-etal-2020-optimus}. \name{} 
uses \underline{bi}directional LSTM as the encoder, LSTM as the decoder, and applies a \underline{mean} pooling layer to the BiLSTM layer to obtain the latent vector. \textsc{Optimus}~\cite{li-etal-2020-optimus} is a Transformer-based VAE model that uses BERT~\cite{devlin-etal-2019-bert} as the encoder and GPT-2~\cite{radford2019language} as the decoder.
Unlike MeanSum~\cite{Chu:2019:MeanSum},
both \name{} and Optimus do not use any additional objectives but the basic VAE objective (i.e., the reconstruction loss with KL regularization):%
\setlength{\abovedisplayskip}{2mm}
\setlength{\belowdisplayskip}{2mm}
\begin{align*}
    \mathcal{L}(\theta, \phi) &= \mathcal{L}_{\text{rec}}+ \beta \mathcal{L}_{KL} \\
    \mathcal{L}_{\text{rec}}(\theta, \phi) &= - \mathbb{E}_{q_{\phi}(\boldsymbol{z}|\boldsymbol{x})}[\log p_{\theta}(\boldsymbol{x}|\boldsymbol{z})]\\
    \mathcal{L}_{KL}(\phi) &= D_{KL}(q_{\phi}(\boldsymbol{z}| \boldsymbol{x}) \| p_{\theta}(\boldsymbol{z})),
\end{align*}
where $\phi$ and $\theta$ are the parameters of the encoder $E$ and decoder $G$.
$\beta$ is a hyper-parameter that controls the strength of the KL regularization $\mathcal{L}_{KL}(\phi)$.
We choose the standard Gaussian distribution $\mathcal{N}(\boldsymbol{0}, \boldsymbol{I})$ as the prior distribution $p_\theta(\boldsymbol{z})$.

\section{Revisiting Simple Average Approach}\label{sec:revisiting}

In this section, we revisit the commonly used simple average approach (\simpleavg) and examine the relations between the aggregated latent vector and the quality of generated summaries.

Taking a simple average is an intuitive way to optimize the aggregated vector in the latent space since it minimizes the total distance between input latent vectors and the aggregated vector. Thus, it appears to be a reasonable design choice and has been adopted by multiple unsupervised opinion summarization models as de-facto standard. 

However, we find that only considering the total distance between the input and the aggregated latent vectors does not always render high-quality summaries. This is because \simpleavg{} is completely ignorant of the decoder performance and the resulting generation. In fact, we observe that \simpleavg{} tends to produce overly generic summaries (as shown in Figure~\ref{fig:overview}), which we refer to as {\em summary vector degeneration}.

To gain a better understanding of the summary vector degeneration problem, we conducted further analysis and discovered two factors that cause this problem: 
(1) simply averaging input latent vectors causes $L_2$-norm shrinkage, and (2) latent vectors with smaller $L_2$-norm tend to be decoded into less informative generations.

\begin{figure}[t]
    \centering
    \includegraphics[width=\linewidth]{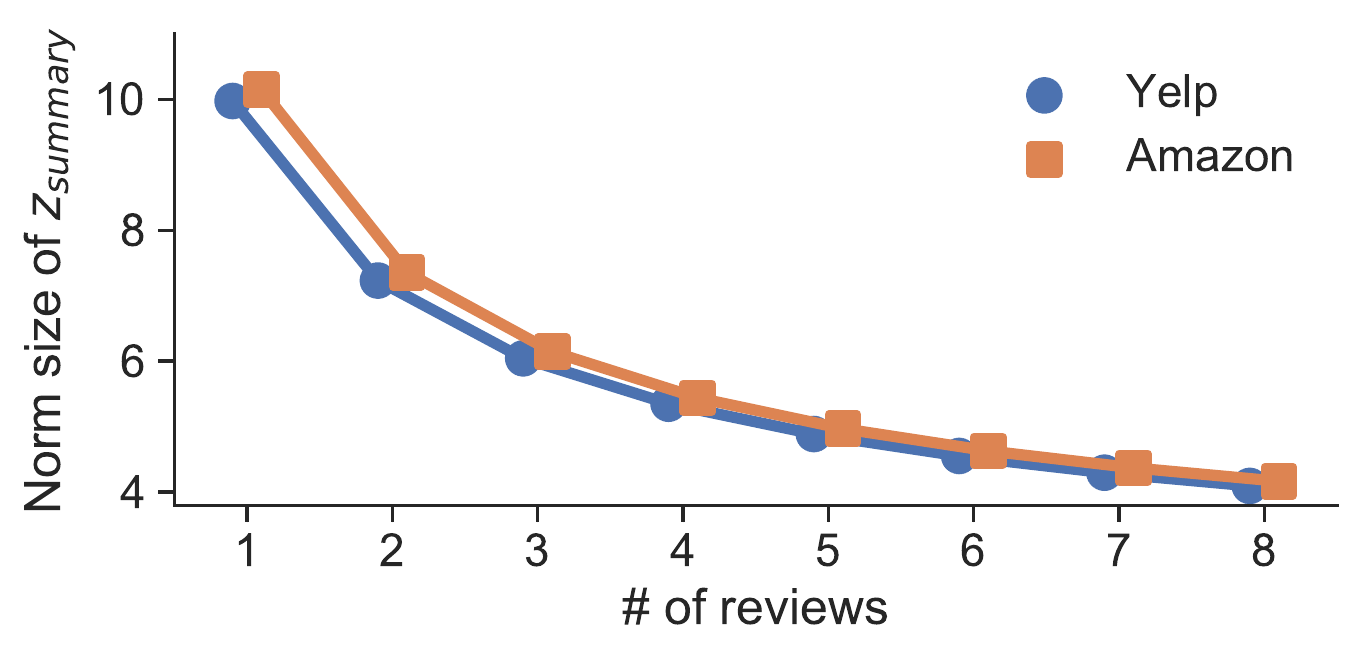}
    \caption{Average $L_2$-norm of simply averaged summary vectors for different number of input reviews.}\label{fig:num_input_vs_norm}
\end{figure}

\subsection{$L_2$-norm Shrinkage in Latent Space}
\label{sub:avg_degenerate}

To understand how simply averaged latent vectors distribute in the latent space, we compared the $L_2$-norm of the latent vectors of input reviews and summary vectors created by \simpleavg.
We conducted analysis using \name{} on two review datasets, \yelp{} and \amazon. %

As shown in Figure~\ref{fig:num_input_vs_norm}, the average $L_2$-norm of the summary vectors significantly {\em shrinks} from 9.97 to 4.10 on \yelp{} (10.15 to 4.17 on \amazon) as the number of input reviews is increased from 1 (i.e., individual reviews) to 8.
The results show that simply averaging multiple latent vectors can cause $L_2$-norm shrinkage of the summary vector.
As we expect each dimension in the latent space to represent a distinct semantics, 
$L_2$-norm shrinkage may cause some information loss in the summary vector.

\subsection{Summary Vector Degeneration}
\label{sub:embed_center}

To investigate the effect of $L_2$-norm shrinkage in the latent space, we further analyzed the quality of generated text for each latent vector and conducted correlation analysis against the $L_2$-norm. 
We used two metrics to assess the quality of generated text: (a) text length and (b) information amount. For the information amount, we trained an autoregressive model (RNN-LM) on each dataset and used negative log probabilities of generated summaries (i.e., a higher value means more amount of information)~\cite{brown-etal-1992-estimate,mielke-etal-2019-kind}\footnote{Training details are in Appendix.}.

Figure~\ref{fig:z_analysis} shows that the $L_2$-norm of latent vectors is highly correlated with (a) generated text length and (b) information amount.
The results support that latent vectors with smaller (larger) $L_2$-norm are decoded into less (more) informative text.
Therefore, we confirm that the commonly used \simpleavg{} is a suboptimal solution for latent vector aggregation as it tends to cause
summary vector degeneration.

\begin{figure}[t]
    \centering
    \begin{subfigure}[b]{\linewidth}
        \centering
        \includegraphics[width=\textwidth]{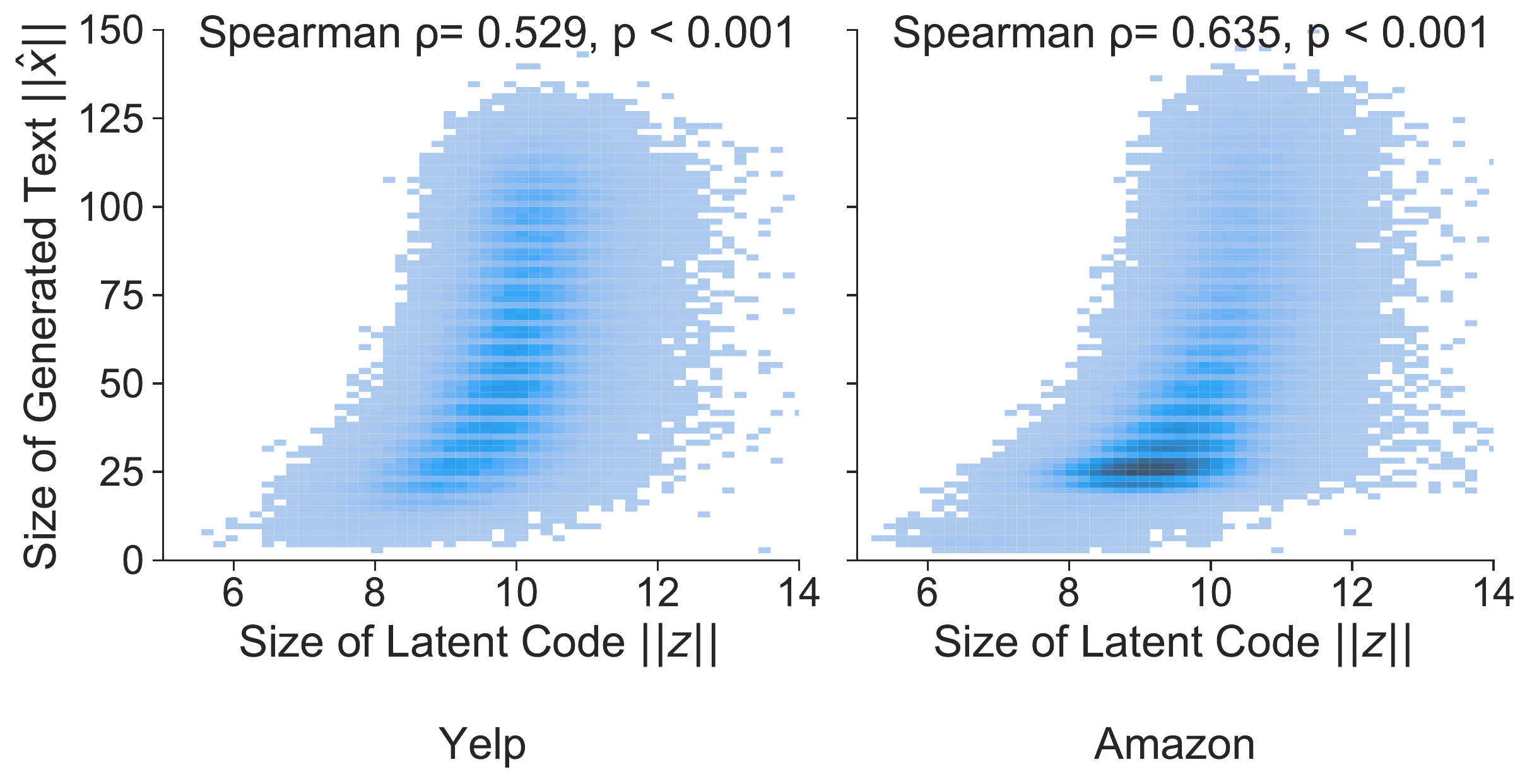}
        \caption{$L_2$-norm vs. generated text length}
        \label{fig:z_x}
    \end{subfigure}
    ~\hfill
    \begin{subfigure}[b]{\linewidth}
        \centering
        \includegraphics[width=\textwidth]{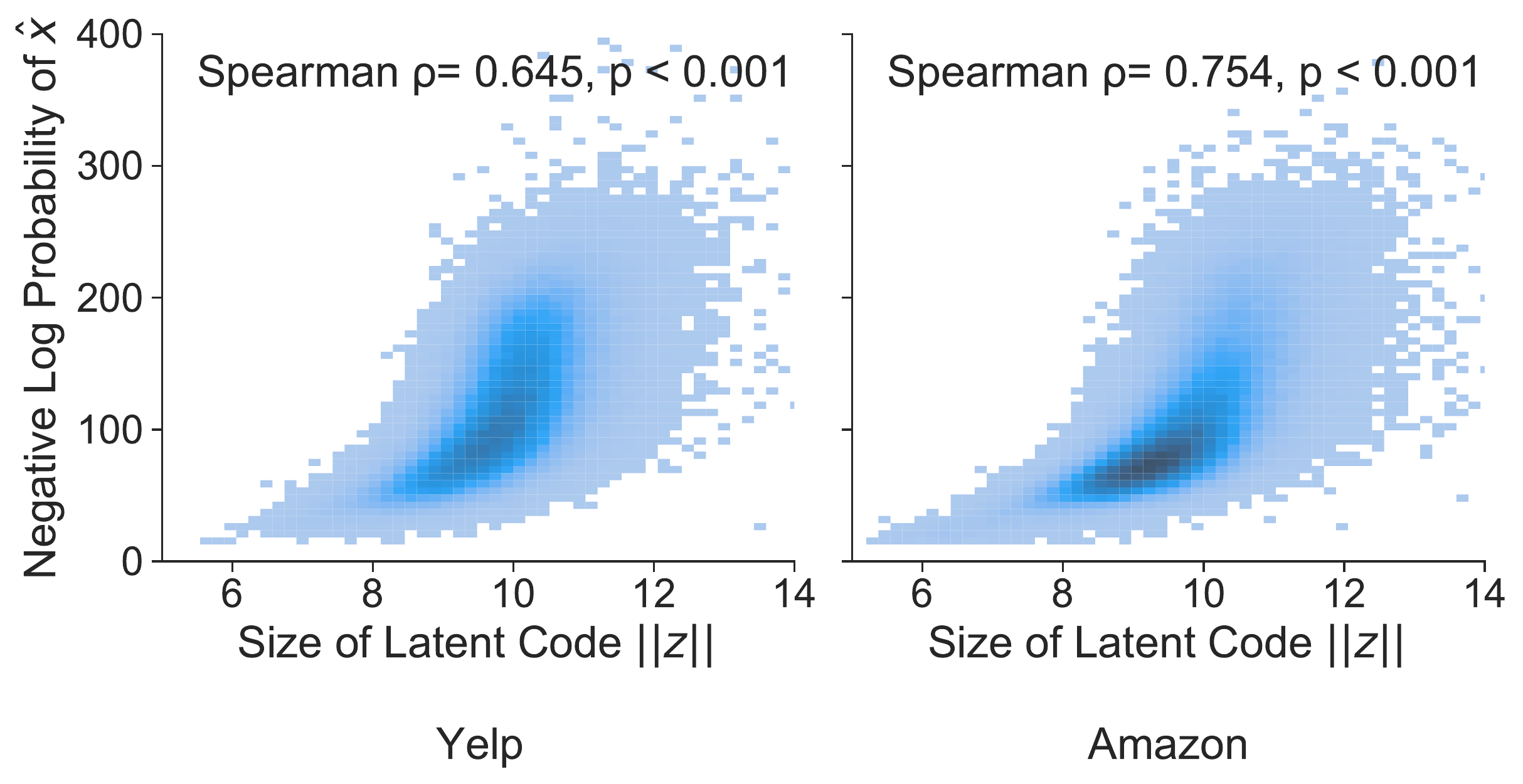}
        \caption{$L_2$-norm vs. information amount of generated text}
        \label{fig:z_prob}
    \end{subfigure}
    \caption{Correlation analysis of the $L_2$-norm of latent vectors $\|z\|$ and the generated text quality: (a) text length and (b) information amount.}
    \label{fig:z_analysis}
\end{figure}

\section{Convex Aggregation in Latent Space}\label{sec:agg}
\hl{As discussed above, there are two limitations} with the de-facto standard \simpleavg. First, it causes summary vector degeneration. Second, it is ignorant of the decoder generation (in the text space $\mathcal{X}$) for a summary vector (in the latent space $\mathcal{Z}$.)

To address the issues, we consider an optimization problem that searches for the best combination of the latent vectors of input reviews that maximizes the {\em alignment} between input reviews and generated summaries.
We restrict the search space to the convex combinations of input review representations, so the contribution of each input review is always zero or positive. This is based on the assumption that each review in the input set should be either ignored or reflected. Hence, we refer to the latent representation aggregation problem as {\em convex aggregation}.

\subsection{\aggname{}: \underline{Co}nvex Aggregation for \underline{Op}inion Summarization}
\begin{figure}[t]
    \centering
    \includegraphics[width=\linewidth]{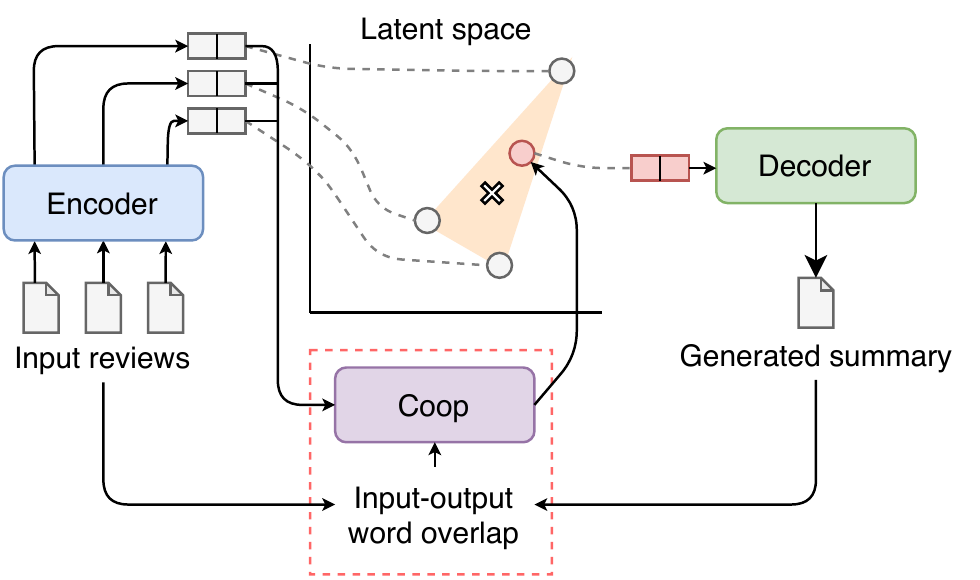}
    \caption{%
    \aggname{} \hl{searches} convex combinations of the latent \hl{vectors} of input reviews based on the {\em input-output word overlap} between a generated summary and input reviews. ${\bf \times}$ denotes the simply averaged vector.}
    \label{fig:coop}
\end{figure}

We develop a latent vector aggregation framework \aggname{} to solve the convex aggregation problem \hl{in Figure~\ref{fig:coop}}. \aggname{} optimizes for the {\em input-output word overlap} between a generated summary and the input reviews:
{
\setlength{\abovedisplayskip}{2mm}
\setlength{\belowdisplayskip}{2mm}
\setlength{\abovedisplayshortskip}{0pt}
\setlength{\belowdisplayshortskip}{0pt}
\begin{align*}
& \underset{\boldsymbol{z}}{\text{maximize}} & & \texttt{Overlap}(\mathcal{R}_e, G(\boldsymbol{z})) \\
& \text{subject to} & & \boldsymbol{z} = \sum_{i = 1}^{|\mathcal{R}_e|} w_i \boldsymbol{z}_i \\
& & & \sum_{i = 1}^{|\mathcal{R}_e|} w_i = 1, \forall w_i \in \mathbb{R}^{+}.
\end{align*}
}

The input-output word overlap ({\tt Overlap}) evaluates the consistency between input reviews and a generated summary, and it can naturally penalize {\em hallucinated generations}. Specifically, we use the ROUGE-1 F1 score as the input-output word overlap metric~\cite{lin-2004-rouge}\footnote{We also tested other ROUGE scores such as ROUGE-2/L in the preliminary experiments and found that ROUGE-1 (i.e., word overlap) works most robustly, so we decided to use the most straightforward metric.}.
Note that the method does not use gold-standard summaries or any information in the test set but uses the input reviews for calculating word overlap information.

\subsection{Search Space}
\label{sub:powerset}
Following the intuition that some input reviews are useful and others are not, we narrow down the search space to the power set of an input review set $R_e$. The summary vector is then calculated as the average of the latent representations of the ``selected'' reviews:
$\boldsymbol{z}_{\texttt{summary}}^{\texttt{power}} = \frac{1}{|\mathcal{R}_e'|} \sum_{i = 1}^{|\mathcal{R}_e'|} \boldsymbol{z}_i,$
where $\mathcal{R}_e' \in 2^{\mathcal{R}_e}\setminus \{\emptyset\}$ is non-empty \hla{subsets} in the power set $2^{\mathcal{R}_e}$.
We also tested black-box 
optimization techniques~\cite{audet2017derivative}
to search the entire \hl{continuous} space, but we did not observe improvements despite the extra optimization cost.

\section{Evaluation}\label{sec:exp}
\noindent
{\bf Dataset:} For our experiments, we used two publicly available datasets, \yelp~\cite{Chu:2019:MeanSum} and \amazon~\cite{brazinskas-etal-2020-unsupervised}. Besides reviews used for training, these two datasets also contain gold-standard summaries for 200 and 60 sampled entities, respectively. For both datasets, the summaries are manually created from 8 input reviews, and we used the same dev/test split, 100/100 for \yelp\ and 28/32 for \amazon, released by their authors for our experiments. 

\noindent
{\bf Experimental settings:}
We used Adam optimizer~\cite{kingma2015adam} with a linear scheduler, whose initial learning rate is set to $10^{-3}$ ($10^{-5}$) for \name{} (Optimus.)
To mitigate the KL vanishing issue, we also applied KL annealing during the training~\cite{kingma2016improved,fu-etal-2019-cyclical,li-etal-2019-surprisingly}.

For generation, we used beam search with a size of 4.
In order to generate summary-like texts, we introduce a technique, \textit{first-person pronoun blocking}, that prohibits to generate first-person pronouns (e.g., I, my, me) during summary generations.
We report the ROUGE-1/2/L F1 scores for the automatic evaluation~\cite{lin-2004-rouge}\footnote{\url{https://pypi.org/project/py-rouge/}}.

\noindent
{\bf Comparative methods:}
We compared our models (i.e., \name{} and Optimus with \aggname) against state-of-the-art opinion summarization models that use \simpleavg{} for latent vector aggregation, namely
TextVAE~\cite{bowman-etal-2016-generating},
MeanSum~\cite{Chu:2019:MeanSum}.
We also coupled \name{} and Optimus with \simpleavg{} to verify the effectiveness of \aggname.
\hl{In addition, we report the performance of other abstractive, extractive or weakly-supervised opinion summarization models.}

Besides the unsupervised summarization models, we also report \hl{two types of oracle methods.}

\noindent
{\bf Oracle (single)}: This oracle method selects a single input review that takes the highest ROUGE-1 F1 score on the gold-standard summary.

\noindent
{\bf Oracle (comb.)}: This oracle method selects the best set of input reviews from the power set $2^{\mathcal{R}_e}\setminus \{\emptyset\}$ of input review set $\mathcal{R}_e$ so that it achieves the highest ROUGE-1 F1 score on the gold-standard summary when \name{} is used as the summarization model. This can also be interpreted as the upper-bound performance of \name{}.

\hl{More details about our evaluation can be found in the Appendix.}
\begin{table*}[t]
\centering
\small
\begin{tabular}{lccccccc}\toprule
& \multicolumn{3}{c}{\yelp} & \multicolumn{3}{c}{\amazon} & \\
\textbf{Method} & \textbf{R1} & \textbf{R2} & \textbf{RL} & \textbf{R1} & \textbf{R2} & \textbf{RL} & \#\textbf{Param} \\\midrule
\aggname\\
\quad \name &\textbf{35.37} & \textbf{7.35} & \textbf{19.94} & \textbf{36.57} & \textbf{7.23} & \textbf{21.24} & 13M \\
\quad Optimus & 33.68 & 7.00 & 18.95 & \underline{35.32} & 6.22 & 19.84 & 239M \\\midrule
\simpleavg \\
\quad \name & 32.87 & 6.93 & \underline{19.89} & 33.60 & 6.64 & \underline{20.87} & 13M \\
\quad Optimus & 31.23 & 6.48 & 18.27 & 33.54 & 6.18 & 19.34 & 239M \\
\quad TextVAE$^\dagger$ & 25.42 & 3.11 & 15.04 & 22.87 & 2.75 & 14.46 & 13M\\
\quad MeanSum$^\dagger$ & 28.46 & 3.66 & 15.57 & 29.20 & 4.70 & 18.15 & 28M\\\midrule
\textit{Abstractive}\\
\quad CopyCat$^\dagger$ & 29.47 & 5.26 & 18.09 & 31.97 & 5.81 & 20.16 & -- \\
\quad Opinosis$^\dagger$ & 24.88 & 2.78 & 14.09 & 28.42 & 4.57 & 15.50 & --\\
\quad DenoiseSum$^\ddagger$ & 30.14 & 4.99 & 17.65 & -- & -- & -- & -- \\

\textit{Extractive}\\
\quad LexRank$^\dagger$ & 25.01 & 3.62 & 14.67 & 28.74 & 5.47 & 16.75 & -- \\
\quad Spectral-BERT$^\flat$ & 30.20 & 4.50 & 17.20 & -- & -- & -- & --\\
\quad QT$^\sharp$ & 28.40 & 3.97 & 15.27 & 34.04 & \underline{7.03} & 18.08 \\
\midrule
\textit{Weakly Supervised}\\
\quad PlanSum$^\ddagger$ & \underline{34.79} & \underline{7.01} & 19.74 & 32.87 & 6.12 & 19.05\\
\quad OpinionDigest$^\natural$ & 29.30 & 5.77 & 18.56 & -- & -- & -- & -- \\
\midrule
\textit{Oracle}\\
\quad single & 31.73 & 4.94 & 16.95 & 35.44 & 7.71 & 20.74 & -- \\
\quad comb. & 42.72 & 10.21 & 24.00 & 40.55 & 8.77 & 23.33 & -- \\
\bottomrule
\end{tabular}
\caption{ROUGE scores on the benchmarks. Bold-faced and underlined denote the best and second-best scores respectively. \aggname{} significantly improves the performance of two summarization models, \name{} and Optimus, and achieves new state-of-the-art performance on both of the benchmark datasets.
$\dagger$ means the results
are copied from \citet{brazinskas-etal-2020-unsupervised}, $\ddagger$ from  \citet{amplayo2021unsupervised}, $\flat$ from 
\citet{wang-etal-2020-spectral}, $\sharp$ from \citet{angelidis2021extractive}, and $\natural$ from \citet{suhara-etal-2020-opiniondigest}. Note that this study classifies OpinionDigest and PlanSum as weakly-supervised summarizers since they use additional information other than review text.
}\label{tab:main_results}
\end{table*}

\subsection{Automatic Evaluation}
As shown in Table~\ref{tab:main_results}, \aggname{} is able to improve both summarization models, \name{} and Optimus, by a large margin. With \aggname{}, \name{} and Optimus obtain the new state-of-the-art performance on both benchmark datasets.
Besides the summarization performance, we also show the model sizes in Table~\ref{tab:main_results}. Note that \name{} performs competitively well against Optimus, which is trained on top of large pre-trained language models and has approximately 20x more parameters than \name. We believe this is due to the simple yet important configuration in the model architecture, which uses a BiLSTM encoder (vs. unidirectional LSTM in TextVAE) and a mean-pooling layer (vs. last hidden state).

Meanwhile, \name{} and Optimus with \aggname{} outperforms Oracle (single), which selects the single review that takes the highest ROUGE score. The results indicate that our aggregation framework takes the quality of unsupervised multi-document opinion summarization to the next stage.

It is worthwhile to note that both VAE variations with the conventional simple average aggregation competitively perform well against the state-of-the-art performance on opinion summarization benchmarks as shown in Table~\ref{tab:main_results}. In contrast to previous study~\cite{brazinskas-etal-2020-unsupervised}, which showed that text VAE performs poorly on the opinion summarization, our modified configuration makes \name{} a competitive baseline for the task.%

\subsection{Human Evaluation}
\begin{table}[t]
\small
    \centering
    \begin{tabular}{l|c|ccc}
    \toprule
    &\multirow{2}{*}{{\bf Info}} & \multicolumn{3}{c}{\bf Content Support} \\
    &  & Fully & Partially & No \\ \midrule
    \aggname{} & {\bf 28.0} & {\bf 38.1\%}	& {\bf 35.7\%} & {\bf 26.2\%}\\
    \simpleavg & 18.0 & 35.4\% & 35.2\% & 29.4\%\\
    CopyCat & -52.0 & 37.6\% & 34.2\% & 28.2\%\\
    PlanSum & 6.0 & 30.7\% &	36.2\% & 33.1\% \\
    \bottomrule
    \end{tabular}
    \caption{Human evaluation on \yelp{} dataset. \aggname{} outperforms the other baseline models on both informativeness (Info) and content support. %
    }
    \label{tab:human}
\end{table}
We conducted human evaluation to assess the quality of generated summaries. More specifically, we collected the generated summaries for entities in the \yelp{} test set with four different models,  \aggname{} (\name), \simpleavg{} (\name), CopyCat and PlanSum. Then, we asked three human judges to evaluate the summaries with two criteria: {\em informativeness} and {\em content support}. 

We first %
asked
human judges to evaluate the informativeness of the generated summaries by the Best-Worst Scaling~\cite{louviere2015best}, which scores each summarization method
with values ranging from -100 (unanimously worst) to +100 (unanimously best). We then asked %
human judges to evaluate the content support of the generated summaries. For each sentence in the generated summary, the judges chose an option from (a) fully supported, (b) partially supported, or (c) not.

We present the human evaluation results in Table~\ref{tab:human}. As shown, summaries generated by \aggname{} are more informative than \simpleavg{}\footnote{\name{} shows robust performance even combined with \simpleavg. 
}and the other baseline models. Meanwhile, \aggname{} also behaves well on content support 
as it generates more sentences with full/partial content support than the other methods.
These results indicate that \aggname{} is able to generate more informative and less hallucinated summaries.
Combined with the automatic evaluation results, we conclude that \aggname{} meaningfully improves the quality of summarization generation.

\section{Analysis}\label{sec:exp:analysis}
\hl{In this section, we conduct a series of additional analysis to verify the effectiveness and efficiency of \aggname{}. We also provide detailed descriptions of the setups and additional analysis in the Appendix.}

\begin{figure}[t]
    \centering
    \includegraphics[width=\linewidth]{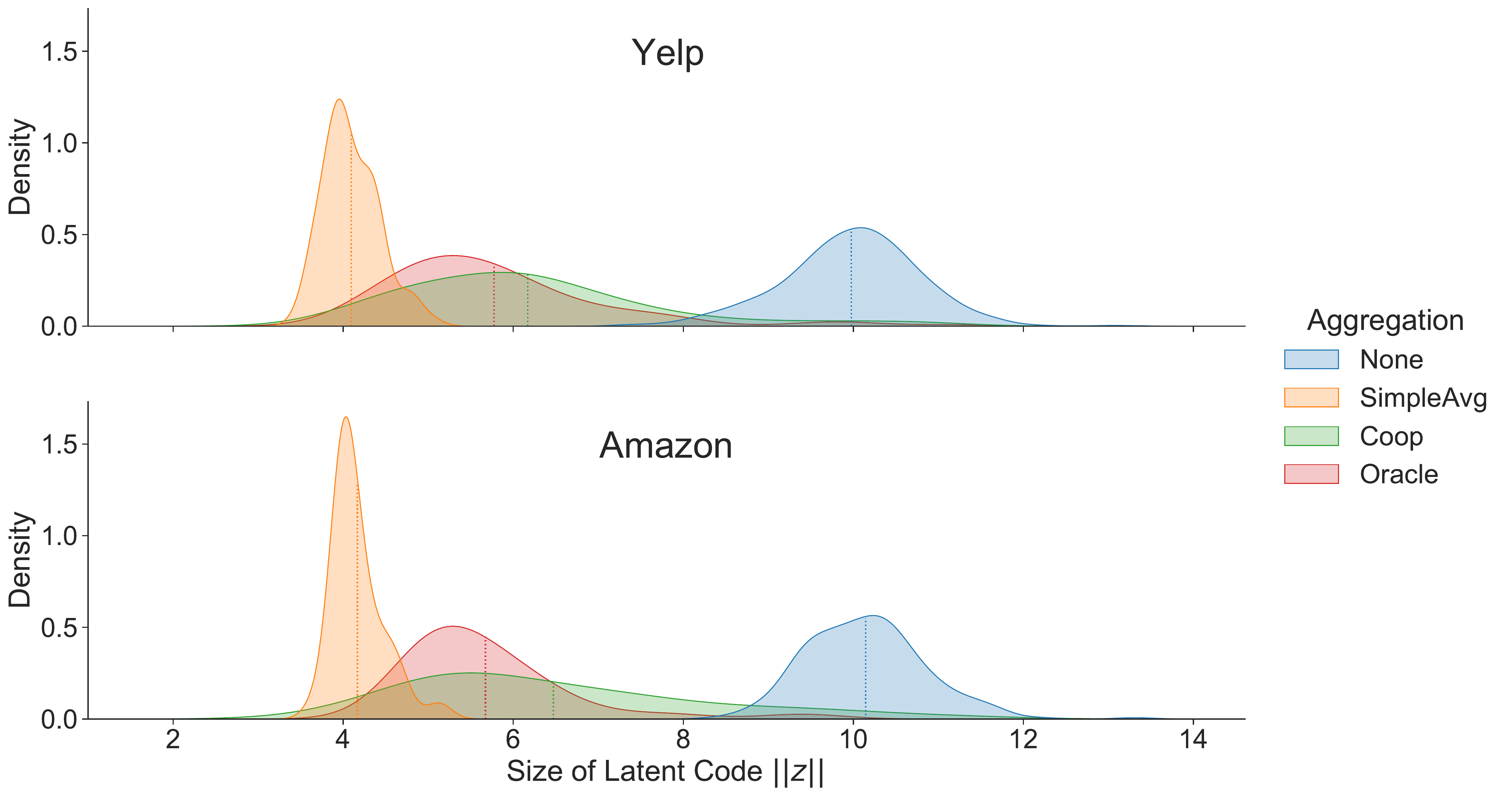}
    \caption{\hl{$L_2$-norm} distributions of latent vectors of input reviews and aggregated vectors.%
    } %
    \label{fig:agg_norm}
\end{figure}

\subsection{Summary Vector Analysis}
\label{sub:summary_vec_analysis}
\noindent
{\bf Does \aggname{} avoid $L_2$-norm shrinkage?}
To verify if \aggname{} alleviates the summary vector degeneration issue, we compare the $L_2$-norm distributions of %
the summary vectors by \simpleavg{}, \aggname{}, Oracle (comb.), and the original reviews (None.) We used \name{} for \simpleavg{} and \aggname{} as the base models.

Figure~\ref{fig:agg_norm} shows \aggname{} does not show severe $L_2$-norm shrinkage compared to \simpleavg{}.
However, the distributions of any aggregation methods, including Oracle, show smaller means of $L_2$-norm compared to individual reviews. This is expected, as customer reviews often contain irrelevant (and specific) information that is not suitable for summaries (e.g., personal experience.)
Therefore, {\em just} preserving the $L_2$-norm of input latent vectors does not necessarily lead to high-quality summary vectors. 

We confirm that \aggname{} has a similar distribution to that of Oracle (comb.), which achieves the upper bound performance of \aggname. The results indicate that \aggname{} successfully excludes input reviews that contain too much irrelevant information, so it can create high-quality summary vectors {\em without} accessing any gold-standard summaries.

\noindent
{\bf How good is \aggname's summary vector?}
We verify how good \aggname's summary vector {\em selections} are with respect to summary generation quality.
Specifically, we sorted all summary vector candidates in power set $2^{\mathcal{R}_e}\setminus\{\emptyset\}$ by the ROUGE-1 score using the generated summary and gold-standard summaries. By doing so, we can use the position of \aggname's selection to evaluate the summary vector quality using ranking metrics. 
We iterated the process for each entity $e$ and used two metrics, namely Mean Reciprocal Rank (MRR) and normalized discounted cumulative gain (nDCG)\cite{schutze2008introduction}, for the evaluation.
We conducted the analysis using \name{} on the test set.
We also evaluated random selection and simple average (i.e., selecting all input reviews.)

\begin{table}[t]
    \centering
    \small
    \begin{tabular}{c|cccc}
    \toprule
        & \multicolumn{2}{c}{\yelp{}} & \multicolumn{2}{c}{\amazon{}}\\
        & MRR & nDCG & MRR & nDCG \\\midrule
        Random & 2.40 & 14.17 & 2.40 & 14.17 \\
        \simpleavg & 4.30 & 16.14 & 1.54 & 13.60\\
        \aggname{} & \textbf{12.05} & \textbf{22.83} & \textbf{14.47} & \textbf{25.21} \\
    \bottomrule
    \end{tabular}
    \caption{Quality of summary vectors for different aggregation methods. \hl{Values are in percentage.}}%
    \label{tab:agg_rank}
\end{table}

As shown in Table~\ref{tab:agg_rank}, \aggname's selection is significantly better than that of the other methods on both of the benchmarks. This confirms that \aggname{} can find {\em good} summary vectors that are decoded into high-quality summaries.
According to the MRR values, \aggname{} selects the 7-8th best-ranked combination (out of 255 candidates) on average.%
The summary quality by the simple average is marginally better (worse) than random selection on \yelp{} (\amazon.)
This is aligned with our findings and discussions in \S\ref{sec:revisiting}, and it further clarifies the negative effects of summary vector degeneration caused by \simpleavg.

\begin{figure}[t]
    \centering
    \includegraphics[width=\linewidth]{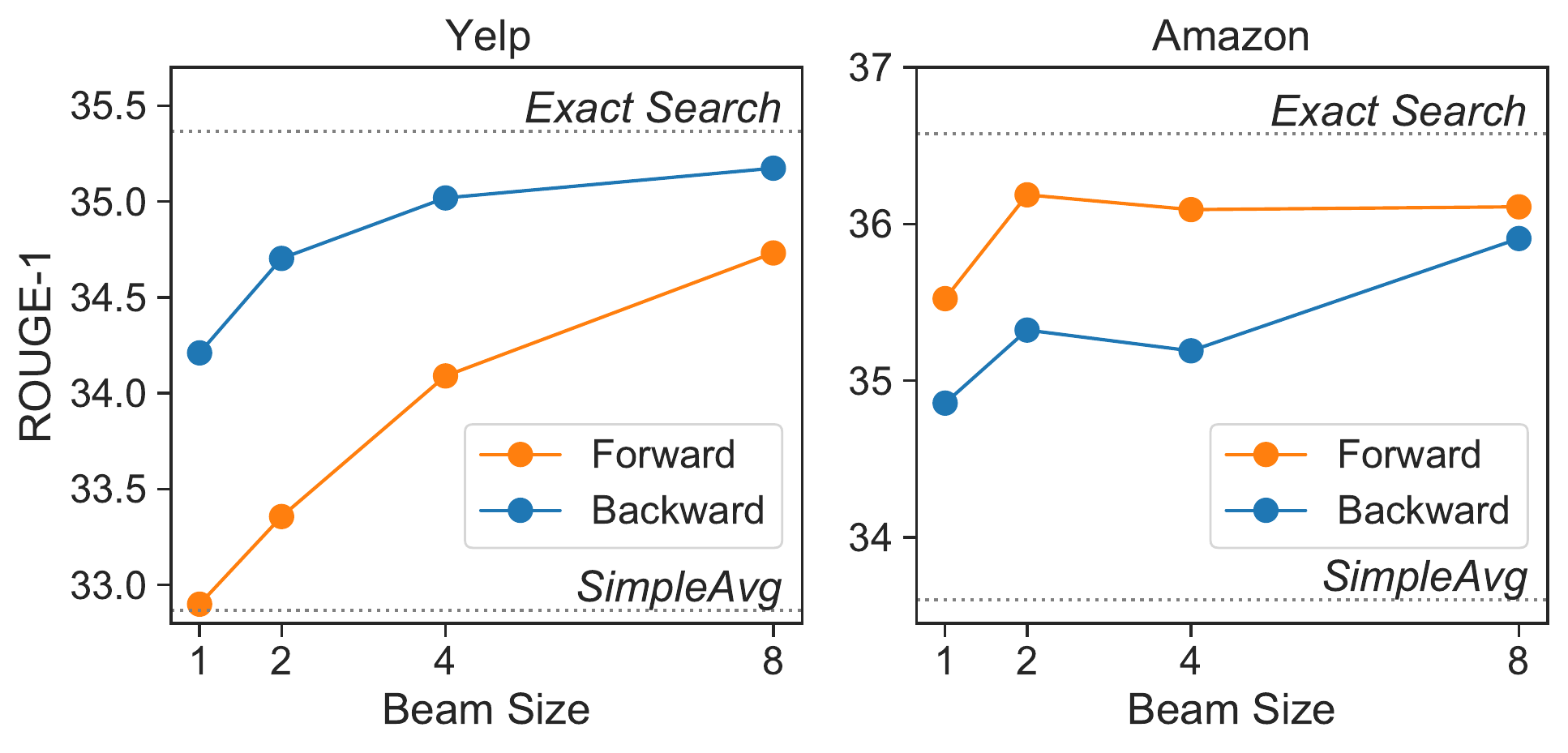}
    \caption{ROUGE-1 scores of \aggname{} with approximate search in different configurations.}%
    \label{fig:beam-1}
\end{figure}

\subsection{Approximate Search}
While we further narrowed down
the search space of \aggname{} into power set in \S\ref{sub:powerset}, the brute-force search becomes intractable for a larger number of input reviews $N$.
Therefore, we tested approximate search algorithms for efficient and effective search.

The simplest solution is the greedy search, which begins from a single review
and progressively adds a review that offers the highest gain in the objective value. The greedy search can be easily generalized to beam search, which stores $k$ candidates for each step.
We also consider the ``inverse'' version of the search algorithms, which begins from all input reviews and removes a review that offers the highest gain by excluding the input review step by step. We refer to the original and the inverse versions as
{\em forward} and {\em backward}, respectively.

Figure~\ref{fig:beam-1} reports the ROUGE-1 performance of \name{} with \aggname{} using approximate search on the \yelp{} and \amazon{} datasets\footnote{ROUGE-2/L are shown
in Appendix.}. The greedy search (beam size = 1) still outperforms the \simpleavg{} baseline, and increasing the beam size further improves the performance of both search methods. 
Thus, we confirm that \aggname{} framework still can provide significant gains by using approximate search when the input size is too large to conduct the exact search for the entire space.

\subsection{Input-output Word Overlap}
To investigate the effectiveness of the input-output word overlap as an intrinsic metric, we analyze the \hl{Spearman's rank} correlation between the input-output word overlap and the ROUGE scores on the gold-standard summaries for each summary vector in the search space $2^{\mathcal{R}_e}\setminus \{\emptyset\}$.

Interestingly, as shown in Table~\ref{tab:correlation}, the two datasets show different trends. In contrast to the Amazon dataset, where the input-output word overlap shows high correlation values against the ROUGE scores, the correlation values on the Yelp dataset are much smaller.
The results confirm that the effect of the input-output word overlap is not just because it is correlated with ROUGE scores between a generated summary and the gold-standard summaries.

\begin{table}[t]
    \centering
    \small
    \begin{tabular}{c|ccc}
    \toprule
      & ROUGE-1 & ROUGE-2 & ROUGE-L \\\midrule
      \yelp & .3758 & .2560 & .2266 \\
      \amazon & .5199 & .3903 & .4816 \\ \bottomrule
    \end{tabular}
    \caption{\hl{Spearman correlation values between the \scorename{} and ROUGE F1 scores on the test set.}}\label{tab:correlation}
\end{table}

\subsection{Qualitative analysis} \label{sec:exp:examples}
Figure~\ref{fig:example-yelp} shows an example of generated summaries using \name{} with the SimpleAvg and \aggname{} for reviews about a restaurant in the \yelp{} dataset.
This example shows that the summary generated from the SimpleAvg $\boldsymbol{z}_{\texttt{summary}}^{\texttt{avg}}$ contains general opinions (e.g., ``the food is delicious.'').
In contrast, \aggname{} effectively chose a subset of reviews to generate a summary vector $\boldsymbol{z}_{\texttt{summary}}^{\texttt{coop}}$, which was decoded into a more specific summary.

\begin{figure*}[th]
    \small
    \vspace{1em}
    \begin{mdframed}
        \textbf{Input Reviews: }\\
        \two{Great service} and clean restaurant. Tonkotsu was excellant. \five{Nice thick broth} and with a little chili oil really hit the spot. Gyoza was excellent and not overfried like some other places. \six{Will return!} </s> \\
        I recommend the hachi special ramen, \five{the broth was delicious} and the noodles cooked just right. We also tried the chashu fried rice which we'll definitely be ordering again. </s>\\
        \gray{This place is great! Small place but so good! The chef taught us about ramen and what he learned from studying ramen in japan! Really interesting! Definitely coming back!!! </s>}
        \\
        \gray{The best ramen in phoenix. They feature tonkotsu, miso and soyu flavored soups and delicious pork in ramen. The owner has trained in japan before coming to arizona and the quality rarely sway dramatically compared to other ramen restaurants when the owner is away. </s>} \\ 
        \gray{Best ramen i've had in phoenix for a very long time. Tradition tonkotsu ramen, shoyu, and a fantastic miso broth are on the menu. The goyza is perfect. </s>}\\ 
        \one{Hachi ramen is delicious}! It is just like being at \three{a small ramen shop} in japan. They focus on \five{their broths creating complex and amazing flavors}. I have tried two of the ramen flavors, their small plates and desserts and have been floored each time. This is the \four{best ramen} in the state and i highly recommend it. </s>\\ 
        \gray{The food here was just ok. The broth was amazing, but my noodles weren't done right. Some were cooked perfectly but others were chewy. Probably will not come back </s>}\\
        \gray{Tonkatsu ramen is the bomb! No msg and broth is so good! The pork is melt in your mouth and not too fatty. The egg has a little infusion of soy, ginger marinade that is extra special! Owner talks to customers and takes great pride! I will be back! I'd take a picture but I ate it too fast! </s> }\\

        \textbf{SimpleAvg} $\boldsymbol{z}_{\texttt{summary}}^{\texttt{avg}}$:\\
        This place is a great place to eat. The \one{food is delicious} and the staff is very friendly. They have a great selection of dishes and the prices are very reasonable. The \two{service is good} and the food is always fresh. It's a great place to go for lunch or dinner.\\
        
        \textbf{\aggname{}} $\boldsymbol{z}_{\texttt{summary}}^{\texttt{coop}}$:\\
        \two{Great service} and \one{delicious food}. It's a \three{\underline{small restaurant}} but the staff is very friendly and attentive. \four{The \underline{ramen} was delicious} and \five{the \underline{broth} was really good}. \six{Will definitely be back.}\\
        \end{mdframed}
    \vspace{-0.5em}
    \caption{Example of summaries generated by \name{} with \simpleavg{} and \aggname{} for reviews about a product on the \yelp{} dataset.
    The colors denote the corresponding opinions, and struck-through reviews in gray were not selected by \aggname{} for summary generation (Note that \simpleavg{} uses all the input reviews.) Terms that are more specific to the entity are underlined. }
    \label{fig:example-yelp}
\end{figure*}

\section{Related Work}
\noindent
{\bf Opinion Summarization~} Multi-document opinion summarization uses the unsupervised approach as it is difficult to collect a sufficient amount of gold-standard summaries for training.
Previously, the common approach was extractive summarization, which selects sentences based on the centrality criteria~\cite{erkan2004lexrank}.
Due to the recent advances in neural network models, unsupervised abstractive summarization techniques have become the mainstream for opinion summarization.

Most abstractive unsupervised opinion summarization techniques use a two-stage approach that trains an encoder-decoder model based on the reconstruction objective and generates a summary from the average latent \hl{vectors} of input reviews using the trained model~\cite{Chu:2019:MeanSum}. %
\citet{amplayo-lapata-2020-unsupervised} and \citet{amplayo2021unsupervised} \hl{have} expanded the approach by creating pseudo review-pairs to train a summarization model.

Our study revisits this two-stage approach and develop a latent \hl{vector} aggregation framework, which can be combined with \hl{a variety of} %
opinion summarization models.

\noindent
{\bf Variational Auto-Encoder~}
The VAE is a variant of auto-encoder that learns a regularized latent space. The text VAE~\cite{bowman-etal-2016-generating}, VAE with autoregressive decoder, has been commonly used for various NLP tasks including text generation~\cite{ye2020variational}, paraphrase generation~\cite{bao-etal-2019-generating} and text style transfer~\cite{hu2017toward,john-etal-2019-disentangled}.

In contrast to the success of the text VAE in NLP tasks, an earlier attempt for using the text VAE for opinion summarization was not successful; \citet{brazinskas-etal-2020-unsupervised} showed that the performance of text VAE was significantly lower than the other baselines.
In this paper, \hl{we devise \name{},} a simple variant of the text VAE\hl{, which} 
performs competitively well against the previous state-of-the-art methods \hl{even when coupled with \simpleavg.}%

Recently, a more expressive text VAE model Optimus~\cite{li-etal-2020-optimus}, which is built on top of pre-trained BERT and GPT-2 models, has been developed. The model was originally developed for sentence generation tasks, and we are the first to combine it with a latent vector aggregation framework for unsupervised opinion summarization tasks.

\section{Conclusions}
In this paper, we revisit the unsupervised opinion summarization architecture and show that the commonly used {\em simple average} aggregation is sub-optimal since it \hl{causes summary vector degeneration and} does not consider the difference in the quality of input reviews or decoder generations.

To address the issues, we develop a latent \hl{vector} aggregation framework \aggname, which searches convex combinations of the latent \hl{vectors} of input reviews based on the word overlap between input reviews and a generated summary. The strategy helps the model generate summaries that are more consistent with input reviews.
To the best of our knowledge, \aggname{} is the first framework that tackles the latent \hl{vector} aggregation problem for opinion summarization.

Our experiments have shown that with \aggname{}, two summarization models, \name{} and Optimus, establish new state-of-the-art performance on two opinion summarization benchmarks. The results demonstrate that our aggregation framework takes the quality of unsupervised opinion summarization to the next stage.

\section*{Acknowledgements}
We thank Arthur Bra{\v{z}}inskas for his valuable feedback and correcting the description about CopyCat.

\bibliography{emnlp2021}
\bibliographystyle{acl_natbib}

\clearpage
\appendix
\section{Dataset Preparation}\label{app:dataset}

\noindent
{\bf Source Dataset:} 
For the \yelp{} dataset, we used reviews provided in the Yelp Open Dataset\footnote{\url{https://www.yelp.com/dataset}}.
For \amazon{} dataset, we used reviews provided in the Amazon product review dataset~\cite{he2016ups}, and we select 4 categories: \textit{Electronics; Clothing, Shoes and Jewelry, Home and Kitchen; Health} and \textit{Personal Care.}

\noindent
{\bf Preprocessing:} We restricted the character set to be ASCII printable for the experiments. We preprocessed the datasets by excluding non-ASCII characters from the reviews and by removing accents from accented characters.

\noindent
{\bf Tokenizer:} 
For \name{}, we used \texttt{sentencepiece}~\cite{kudo-richardson-2018-sentencepiece}\footnote{
\url{https://github.com/google/sentencepiece}} to train a BPE tokenizer with a vocabulary size of 32k, a character coverage of 100\%, and a max sentence length of 8,192.

For Optimus, we used the pre-trained tokenizers provided by \texttt{transformers}~\cite{wolf-etal-2020-transformers}\footnote{\url{https://github.com/huggingface/transformers}}.
Since Optimus uses different models for the encoder and decoder, we used different tokenizers for the encoder ({\tt bert-base-cased}) and the decoder ({\tt gpt2}), respectively.

\noindent
{\bf Training data:} We used pre-defined training sets of \yelp{} and \amazon{} with additional filtering. We filtered out reviews that consist of more than 128 tokens after tokenization by the BPE tokenizer trained for \name. 
As a result, the training sets contain 3.8 million and 13.3 million reviews in \yelp{} and \amazon{} respectively.
We further excluded entities that have less than 10 reviews.
The basic statistics of the training data after those filtering steps are shown in Table~\ref{tab:train_set_stats}.

\begin{table}[ht]
\small
    \centering
    \begin{tabular}{lcc}
    \toprule
         & \yelp & \amazon \\\midrule
\# of entity & 75,988 & 244,652 \\
\# of reviews & 4,658,968 & 13,053,202 \\
\bottomrule
    \end{tabular}
    \caption{Statistics of the filtered training data.}
    \label{tab:train_set_stats}
\end{table}

\section{Revisiting Simple Average Approach}

\subsection{RNN-LM model for information amount}
\label{app:rnnlm}
We trained single-layer RNN-LMs on Yelp and Amazon datasets respectively, in 100k steps with a batch size of 256.
The embedding size and the hidden size are set to 512 and the output vocabulary layer is tied with the input embedding layer~\cite{press-wolf-2017-using,inan2017tying}.

\subsection{Additional analysis on latent vector and input text quality}
\label{app:z_analysis}
In \S 3, we investigated the relationship between $L_2$-norm of latent vectors and the generated text qualities, and found strong positive correlations.
In addition to the generated reviews $\hat{x}$, we conduct the same analysis using input reviews $x$.

\begin{figure}[th]
    \centering
    \begin{subfigure}[b]{0.99\linewidth}
        \centering
        \includegraphics[width=\textwidth]{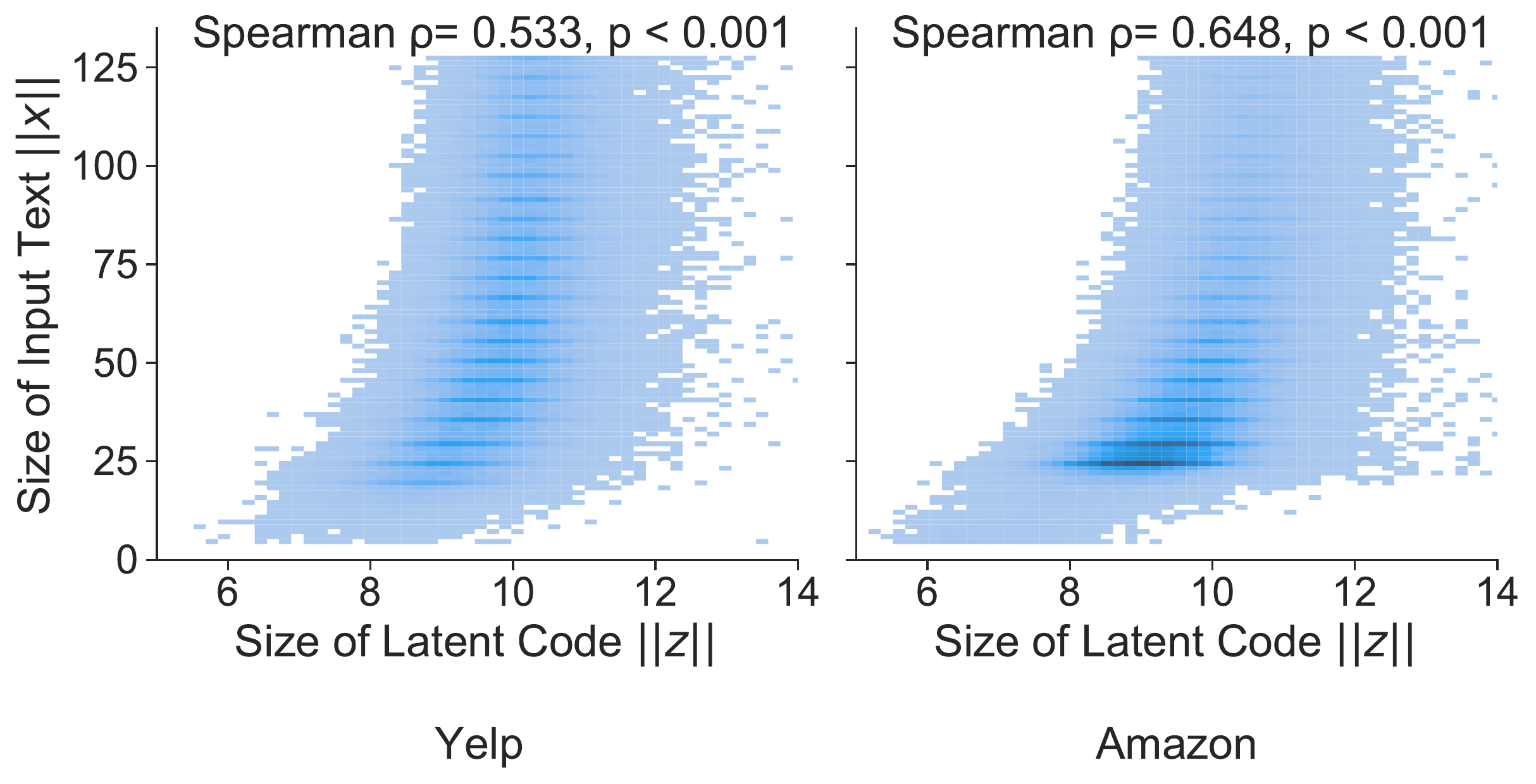}
        \caption{$L_2$-norm vs. input text length}
        \label{fig:z_x_app}
    \end{subfigure}
    ~\hfill
    \begin{subfigure}[b]{0.99\linewidth}
        \centering
        \includegraphics[width=\textwidth]{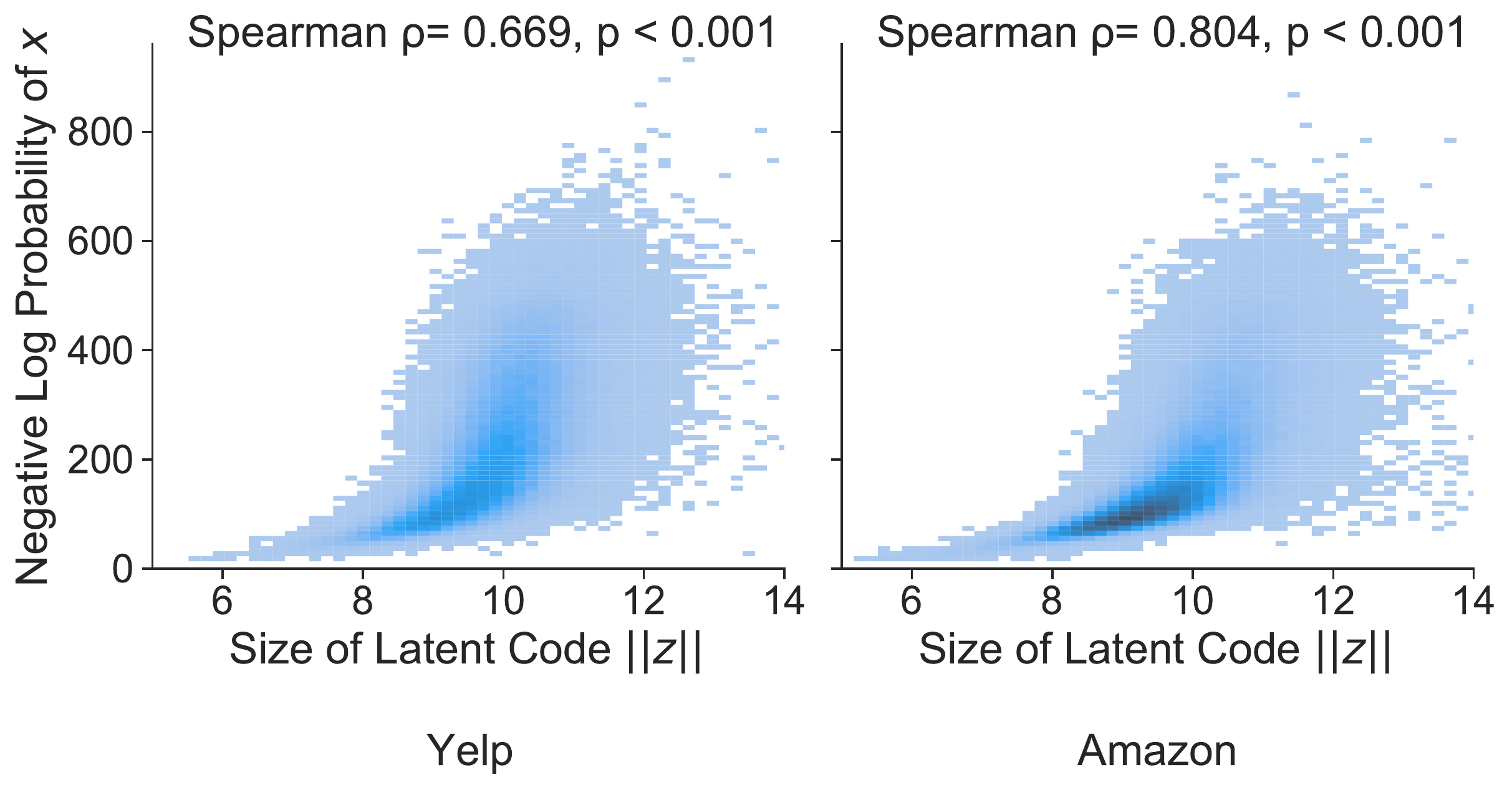}
        \caption{$L_2$-norm vs. information amount of input text}
        \label{fig:z_prob_app}
    \end{subfigure}
    \caption{Illustrations of the relationships between $L_2$-norm of latent vectors $\|z\|$ and the input review quality: (a) text length and (b) the information content.}
    \label{fig:z_analysis_app}
\end{figure}

As shown in Figure~\ref{fig:z_analysis_app}, we confirm the same trends that $L_2$-norm of latent vectors is highly correlated with both metrics. Thus, we confirm that less (more) informative text tends to be embedded closer to (more distant from) the origin in the latent space.

\section{Evaluation}

\begin{table*}[ht]
\centering
\begin{tabular}{lccccccc}\toprule
& \multicolumn{3}{c}{\yelp} & \multicolumn{3}{c}{\amazon} & \\
\textbf{Method} & \textbf{R1} & \textbf{R2} & \textbf{RL} & \textbf{R1} & \textbf{R2} & \textbf{RL} & \#\textbf{Param} \\\midrule
\aggname\\
\quad \name &\textbf{36.16} & 7.25 & \underline{20.09} & \textbf{36.30} & \underline{6.81} & \textbf{21.11} & 13M \\
\quad Optimus & \underline{35.51} & \textbf{7.84} & 19.27 & \underline{35.98} & \textbf{7.17} & \underline{20.16} & 239M \\\midrule
SimpleAvg\\
\quad \name & 33.38 & 7.25 & \textbf{20.32} & 31.80 & 6.04 & 20.00 & 13M \\
\quad Optimus & 33.87 & \underline{7.67} & 18.98 & 33.66 & 6.51 & 19.60 & 239M \\
\midrule
\textit{Oracle}\\
\quad comb. & 44.40 & 11.78 & 24.02 & 39.66 & 8.73 & 22.75 & -- \\
\bottomrule
\end{tabular}
\caption{ROUGE scores on the development set of benchmarks. Bold-faced and underlined denote the best and second-best scores respectively.
}\label{tab:dev_results}
\vspace{-0.2cm}
\end{table*}

\subsection{Training settings}
\label{app:exp_settings}
Major hyper-parameters for training models are reported in Table~\ref{tab:bimeanvae_param} and \ref{tab:optimus_param} following the ``Show-Your-Work'' style suggested by \citet{dodge-etal-2019-show}.

\noindent
{\bf Optimization:} We used Adam optimizer~\cite{kingma2015adam} with a linear scheduler that has warm-up steps. The initial learning rate was set to be $10^{-3}$ for \name{} and $10^{-5}$ for Optimus.

\noindent
{\bf KL annealing:}
For the VAE training, we adopt the KL annealing to avoid the KL vanishing issue~\cite{bowman-etal-2016-generating}.
To be specific, we tested two KL annealing strategies to control the $\beta$ value, i.e., cyclical KL annealing~\cite{fu-etal-2019-cyclical} and pretrain-then-anneal with KL thresholding (aka FreeBits)~\cite{kingma2016improved,li-etal-2019-surprisingly}.
{\bf The cyclic KL annealing} repeats the monotonic annealing process of the $\beta$ parameter from 0 to 1 multiple times~\cite{fu-etal-2019-cyclical}.
{\bf The pretrain-then-annealing approach} has two steps~\cite{li-etal-2019-surprisingly}. The first step pre-trains an autoencoder model with the $\beta$ parameter fixed to 0.
The second step re-initializes the decoder parameter and trains the model with the $\beta$ parameter monotonically increased from 0 to 1.

In addition to the annealing schedule, we also searched a threshold hyper-parameter for the KL value to control the strength of the KL regularization~\cite{kingma2016improved}.

\subsection{Baseline Models}
\label{app:baselines}
We considered multiple abstractive and extractive summarization models. 

\noindent
{\bf \vaeorig}~\cite{bowman-etal-2016-generating}: A vanilla text VAE model that has a unidirectional LSTM layer and uses the last hidden state to calculate the parameters of the posterior distribution. The model was tested in \citet{brazinskas-etal-2020-unsupervised} but performed poorly.

\noindent
{\bf MeanSum} \cite{Chu:2019:MeanSum}: An unsupervised multi-document abstractive summarization method that minimizes a combination of the reconstruction and similarity loss.

\noindent
{\bf CopyCat}~\cite{brazinskas-etal-2020-unsupervised}: An unsupervised opinion summarization model. CopyCat incorporates an additional latent vector $c$ to model an entire review set $R_e$ in addition to latent vectors for individual reviews.
This hierarchical modeling enables CopyCat to consider both global (entity-level) and local (review-level) information to calculate a latent representation.

\subsection{Performance on Development Set}
We report the performance on the development set in Table~\ref{tab:dev_results}.
\aggname{} consistently improve the performance on ROUGE scores (except for ROUGE-L on \yelp) on the development set.

\begin{table*}[ht]
\centering
\begin{tabular}{llcccccc}\toprule
& \multicolumn{3}{c}{\yelp} & \multicolumn{3}{c}{\amazon} \\
\multicolumn{2}{l}{\textbf{Aggregation}} & \multicolumn{1}{c}{\bf R1} & \multicolumn{1}{c}{\bf R2}  &\multicolumn{1}{c}{\bf RL}  & \multicolumn{1}{c}{\bf R1} & \multicolumn{1}{c}{\bf R2}  &\multicolumn{1}{c}{\bf RL}   \\\midrule
\multicolumn{2}{l}{Extractive} & {\color{red} 30.51$\downarrow$} & {\color{red}5.34$\downarrow$} & {\color{red}18.42$\downarrow$} & {\color{red}32.43$\downarrow$} & {\color{red}6.13$\downarrow$} & {\color{red}20.11$\downarrow$} \\
\multicolumn{2}{l}{Inverse-Variance Weighting} & {\color{red}32.15$\downarrow$} & {\color{red}6.63$\downarrow$} & {\color{darkgreen} 20.11$\uparrow$} & {\color{red}33.18$\downarrow$} & {\color{red}6.33$\downarrow$} & {\color{red}20.72$\downarrow$} \\
\multicolumn{2}{l}{Policy Gradient} & {\color{red}32.21$\downarrow$} & {\color{red}6.77$\downarrow$} & {\color{red}19.46$\downarrow$} & {\color{red}33.33$\downarrow$} & {\color{red}6.12$\downarrow$} & {\color{red}20.58$\downarrow$}\\
Rescale & $\alpha = 1$ &  {\color{red}31.63$\downarrow$} &  {\color{red}6.38$\downarrow$} &  {\color{darkgreen} 20.56$\uparrow$} &  {\color{red} 30.11$\downarrow$} & {\color{red} 4.74$\downarrow$} &  {\color{red} 19.45$\downarrow$} \\
& $\alpha =  5$  &  {\color{red} 32.77$\downarrow$} & {\color{red} 6.87$\downarrow$} &  {\color{red}19.57$\downarrow$} &  {\color{darkgreen} 34.01$\uparrow$} & {\color{darkgreen}  6.68$\uparrow$} &  {\color{darkgreen} 21.34$\uparrow$} \\
& $\alpha =  10$ &  {\color{red} 30.38$\downarrow$} &  {\color{red} 5.85$\downarrow$} &  {\color{red} 18.10$\downarrow$} &  {\color{darkgreen} 34.11$\uparrow$} &  {\color{red} 6.55$\downarrow$} &  {\color{red}20.33$\downarrow$} \\\midrule
\multicolumn{2}{l}{SimpleAvg} & 32.87 &  6.93 &  19.89 & 33.60 & 6.64 &  20.87 \\\bottomrule
\end{tabular}
\caption{ROUGE scores of \name{}, coupled with different input aggregation methods.
}\label{tab:other_agg}
\end{table*}

\subsection{Baselines for summary vector degeneration}

In this paper, we develop a latent vector aggregation framework based on the input-output word-overlap to address the summary vector degeneration problem. As alternative and reasonable solutions, we tested the following methods and confirm that none of them consistently outperforms \simpleavg, as shown in Table~\ref{tab:other_agg}.

\paragraph{Extractive} This method uses an extractive summarization technique to select $k$ input reviews that best represent the input review set. In the analysis, we used LexRank and set $k=4$, which was chosen based on the Oracle (comb.) performance on the dev set. We used the simple average for the latent representation aggregation.

\paragraph{Inverse-Variance Weighting}
Weighted average based on importance scores of input reviews is an alternative way for latent representation aggregation.
Specifically, we consider the variance parameter of posterior $\operatorname{diag}\boldsymbol{\sigma}^2$ as the importance score of each input review.
We found that generic reviews (i.e., reviews that do not describe entity-specific information) tend to have  large variance parameters. To reduce the influence by such kind of generic input reviews,
we use the inverse-variance weighting~\cite{cochran1954combination} to assign larger weights to input reviews that contain more entity-specific information:
\[\textstyle
    \boldsymbol{z}_{\texttt{summary}}^{\texttt{ivw}} = \left(\sum_{i = 1}^{N_e} \boldsymbol{\sigma}_{i}^{-1}\right)^{-1} \sum_{i=1}^{N_e} \boldsymbol{\sigma}_{i}^{-1} \boldsymbol{z}_{i}.
\]

\paragraph{Policy Gradient} We also used reinforcement learning to optimize the convex aggregation problem.
In particular, we used the self-critical policy gradient~\citep[PG;][]{rennie2017self, paulus2018deep} to search the weights of input reviews:
\[\textstyle
    \mathcal{L}_{\text{PG}} = (r(y^s)- r(\hat{y}))\sum_{t=1}^{|y^s|}\text{log}~p(y^s_t|y^s_{<t},\mathcal{R}_e)),
\]
where the reward function $r$ is the input-output word overlap described in Section~\ref{sec:agg}. For each entity, we froze the encoder-decoder parameters and trained the input review weights with $\mathcal{L}_{\text{PG}}$ for $10$ epochs with an initial learning rate $10^{-2}$.

\begin{figure}[t]
    \centering
    \begin{subfigure}[b]{0.99\linewidth}
        \centering
        \includegraphics[width=\textwidth]{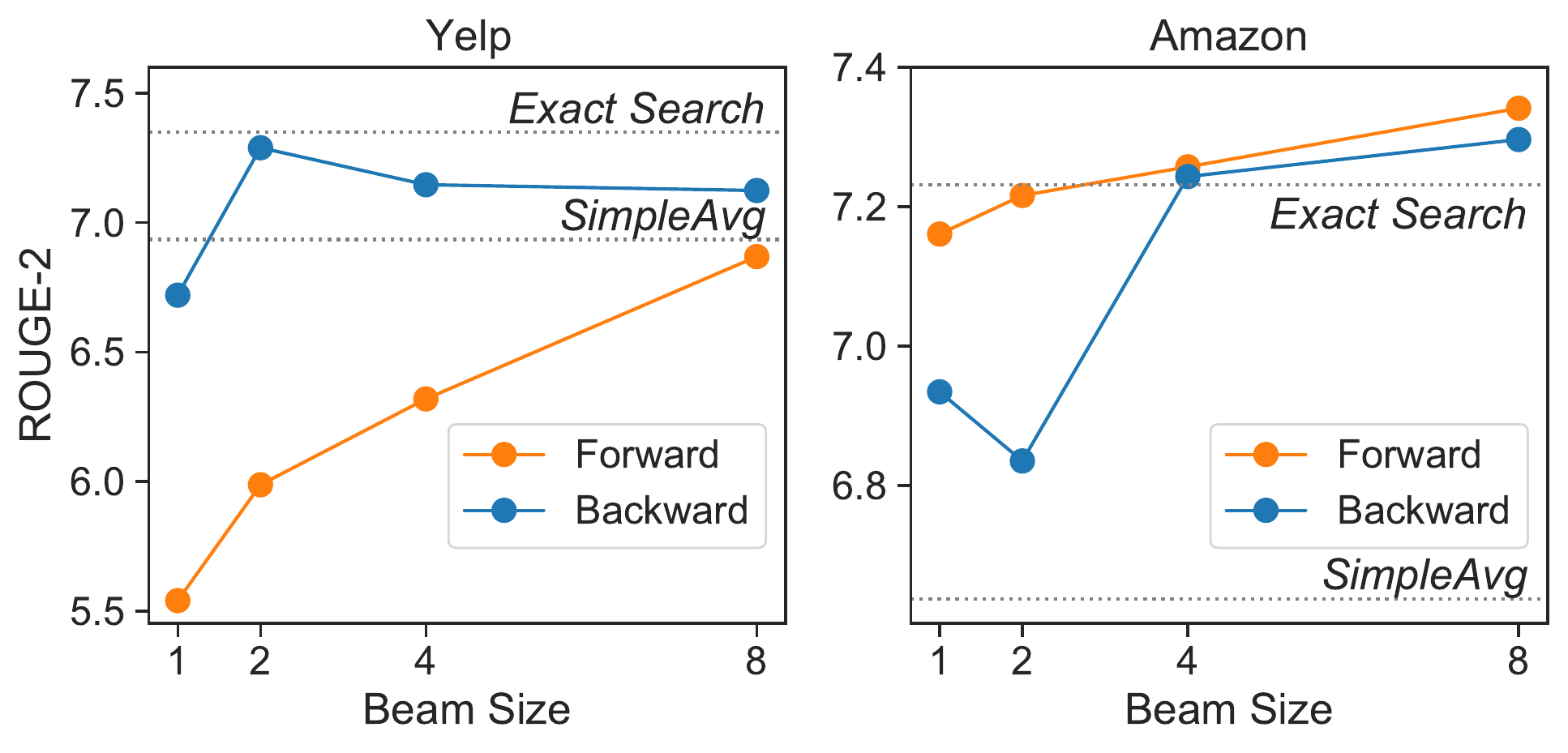}
        \caption{ROUGE-2}
        \label{fig:beam-2}
    \end{subfigure}
    \begin{subfigure}[b]{0.99\linewidth}
        \centering
        \includegraphics[width=\textwidth]{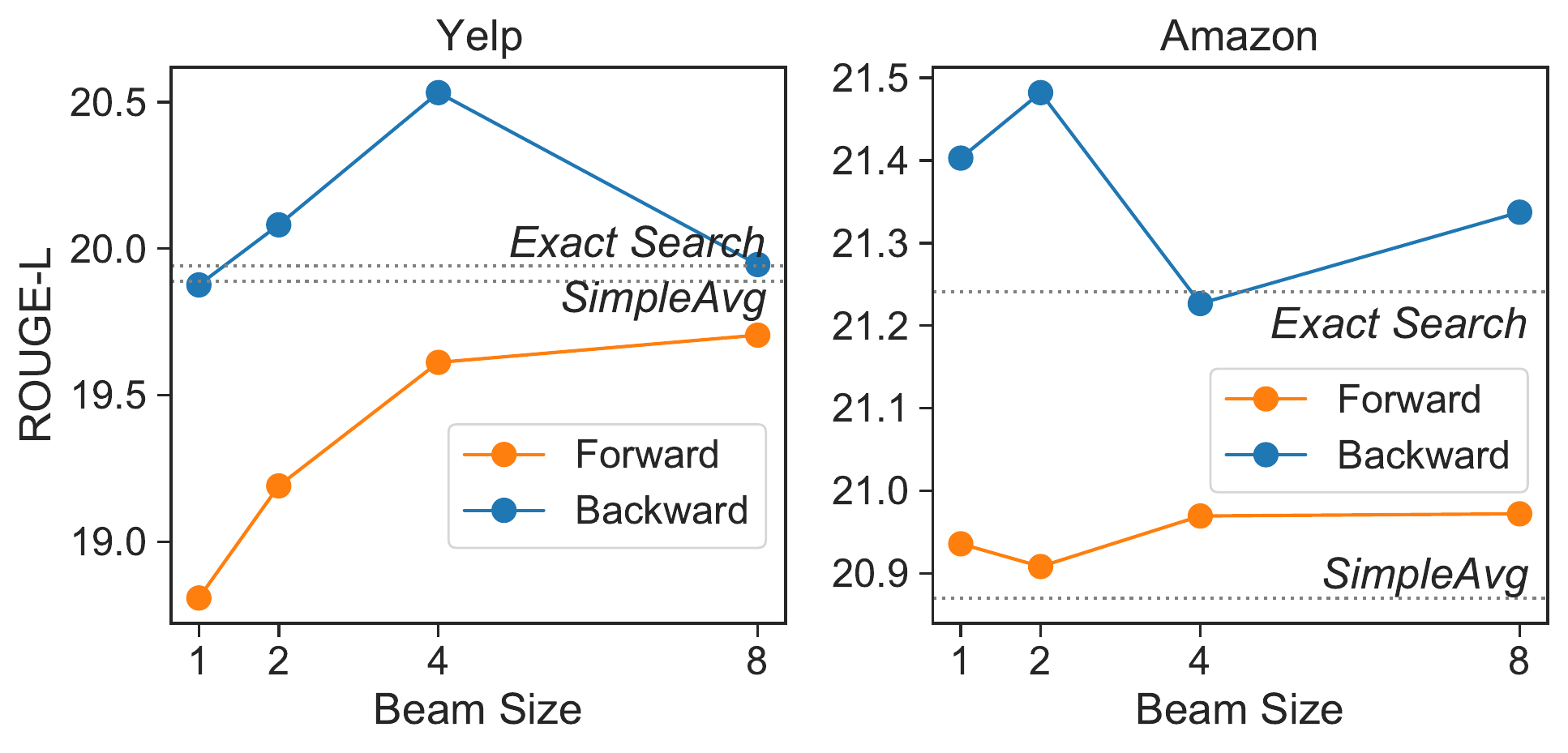}
        \caption{ROUGE-L}
        \label{fig:beam-l}
    \end{subfigure}
    \caption{Approximated search performance of ROUGE-2/L scores with different batch sizes.}
    \label{fig:beam-2l}
\end{figure}

\paragraph{Re-scaling}
The last baseline approach is to re-scale the aggregated latent vector. Specifically, we first normalize the averaged latent vector $\boldsymbol{z}^{\texttt{avg}}_{summary}$ (Section~\ref{sec:preliminary}) and then re-scale the normalized vector with a constant value $\alpha \in \{1, 5, 10\}$:
\begin{align*}
    \boldsymbol{z}^{\texttt{rescale}}_{\texttt{summary}} = \alpha \cdot \frac{\boldsymbol{z}^{\texttt{avg}}_{summary}}{\|\boldsymbol{z}^{\texttt{avg}}_{summary}\|}
\end{align*}

\section{Analysis}
\subsection{Ranking Metrics} %
\label{app:rank_metrics}
We describe the details of the ranking metrics used in \S 6.1.

\noindent
{\bf Mean Reciprocal Rank (MRR):}
\begin{align*}
    \text{MRR} = \frac{1}{|\mathcal{Z}^{\text{test}}_{\texttt{agg}}|} \sum_{\boldsymbol{z} \in \mathcal{Z}^{\text{test}}_{\texttt{agg}}} \frac{1}{\mathtt{rank}(\boldsymbol{z})},
\end{align*}
{\bf Normalized Discounted Cumulative Gain (nDCG):}
\begin{align*}
    \text{nDCG} = \frac{1}{|\mathcal{Z}^{\text{test}}_{\texttt{agg}}|} \sum_{\boldsymbol{z} \in \mathcal{Z}^{\text{test}}_{\texttt{agg}}} \frac{1}{\log_2(\mathtt{rank}(\boldsymbol{z}) + 1)},
\end{align*}
where $\mathcal{Z}^{\text{test}}_{\texttt{agg}}$ is the set of summery vectors for each aggregation method on test set, and $\mathtt{rank}(\boldsymbol{z})$ denotes the rank of the summary vector $\boldsymbol{z}$ selected by the respective aggregation method in the search space $2^{\mathcal{R}_e}\setminus\{\emptyset\}$.

\subsection{Approximate Search}
\label{app:beam-2l}
In addition to the ROUGE-1 scores shown in the main paper, we show the approximated search performance for ROUGE-2/L scores in Figure~\ref{fig:beam-2l}.
We observe that the ROUGE-2 score shows that backward search is better for the \yelp{} dataset and forward search is better for the \amazon{} dataset, similar to the ROUGE-1 score.
In contrast to ROUGE-1/2, the ROUGE-L score shows that backward search is always better.

\subsection{Runtime Analysis}
\label{app:runtime}
\begin{figure}[t]
    \centering
    \includegraphics[width=0.9\linewidth]{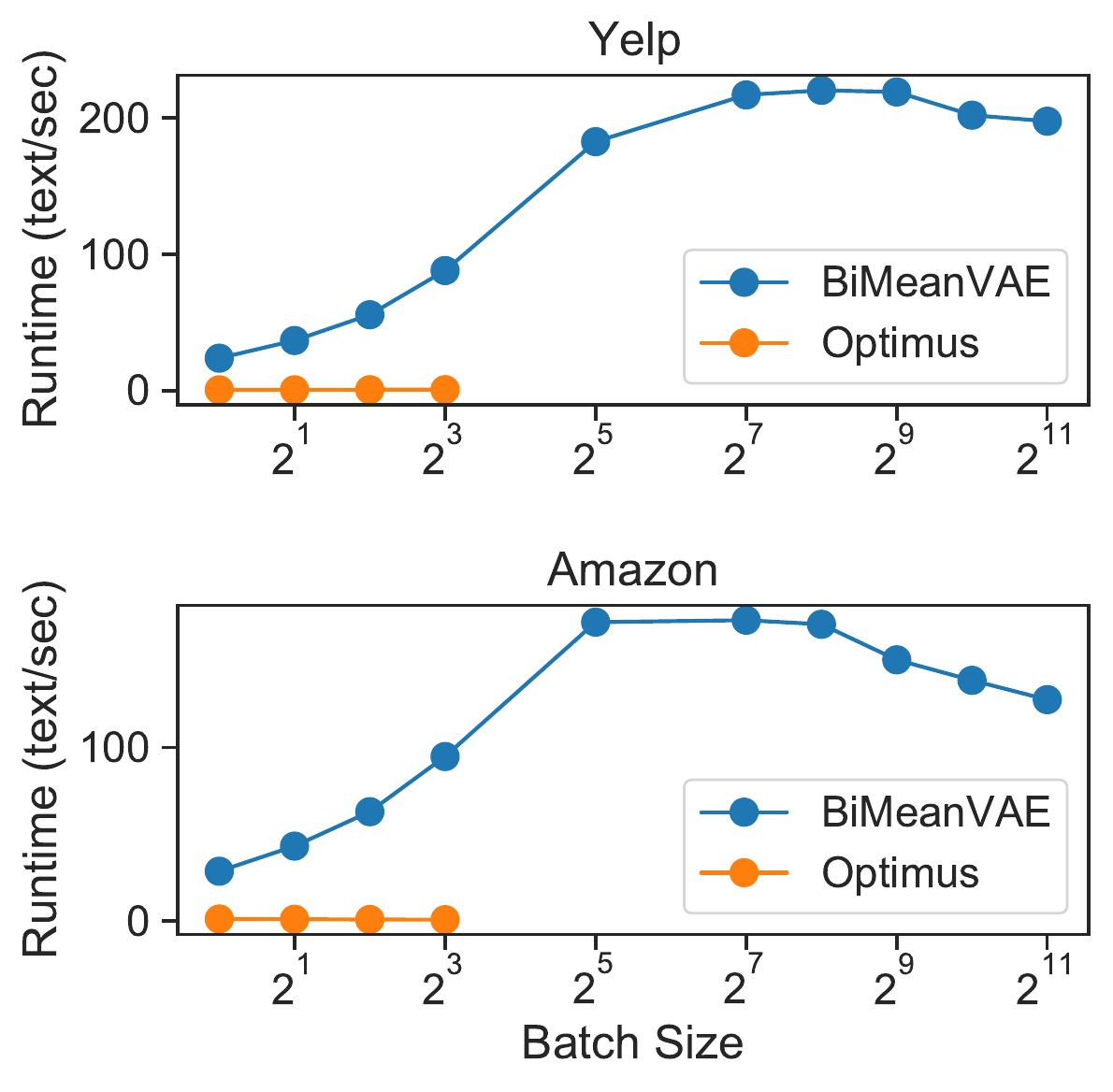}
    \caption{The inference runtime of \name{} and Optimus with different batch sizes.}
    \label{fig:runtime}
\end{figure}
We report the inference runtime of \name{} and Optimus in Figure~\ref{fig:runtime}.
For the \yelp{}/\amazon{} data, \name{} can generate 220.17/173.59 reviews/sec on average, while Optimus can generate only 0.68/1.16 reviews/sec on average.
Optimus is a huge model that uses BERT as the encoder and GPT-2 as the decoder, which makes it more computationally expensive than \name.
Nevertheless, the inference time is still acceptable for running summarizers in batch processing. 

However, due to the GPU memory size limitation, it becomes infeasible for Optimus to take a batch size of more than 8, while \name{} can process much larger batches within a reasonable time.

\begin{figure*}[th]
    \small
    \vspace{1em}
    \begin{mdframed}
        \textbf{Input Reviews: }\\
        I usually wear size 37, but found a 38 feels better in this sandal. I absolutely \four{love this sandal}. So \two{supportive and comfortable}, although at first I did get a blister on my big toe. Do not let this be the deciding factor. It stretched out and is now fabulous. I love it so much that I bought it in three colors. </s> \\
        \gray{This is a really cute shoe that feels very comfortable on my high arches. The strap on the instep fits my feet very well, but I have very slim feet. I can see how it would be uncomfortably tight on anyone with more padding on their feet. </s>}\\
        \gray{I love these sandals. The fit is perfect for my foot, with perfect arch support. I don't think the leather is cheap, and the sandals are very comfortable to walk in. They are very pretty, and pair very well with pants and dresses. </s>}\\
        My wife is a nurse and wears dansko shoes. We were excited to try the new crimson sandal and normally order 39 sandal and 40 closed toe. Some other reviews were right about a  \three{narrow width and tight toe box}. We gave them a try and passed a \one{great pair of shoes} to our daughter with her long narrow feet, and \four{she loves them}... </s>\\
        \gray{Finally, a Dansko sandal that's fashion forward! It was love at first sight! This is my 4th Dansko purchase. Their sizing, quality and comfort is very consistent. I love the stying of this sandal and I'm pleased they are offering bolder colors. Another feature I love is the Dri-Lex topsole - it's soft and keeps feet dry. </s>} \\
        \gray{I really love these sandals. my only issue is after wearing them for a while my feet started to swell as I have a high instep and they were a little tight across the top. I'm sure they will stretch a bit after a few wears </s>}\\
        I have several pairs of Dansko clogs that are all size 39 and fit perfectly. So I felt confident when I ordered the Tasha Sandal in size 39. I don't know if a 40 would be too large but the 39 seems \three{a little small}. Otherwise, \four{I love them}. They are very cushiony and \two{comfortable}! </s>\\
        \gray{I own many Dansko shoes and these are among my favorites. They have ALL the support that Dansko offers in its shoes plus they are very attractive. I love the the heel height and instant comfort. They look great with slacks and dresses, dressed up or not... </s>}\\

        \textbf{SimpleAvg} $\boldsymbol{z}_{\texttt{summary}}^{\texttt{avg}}$:\\
        \one{This is a great shoe}. It is \two{very comfortable}, and the fit is perfect. The only issue is that it's {\color{red}\bf\sout{a little big on the toe area}}, but it's not a problem. It is \two{very comfortable to wear and it's very comfortable}.\\
        
        \textbf{\aggname{}} $\boldsymbol{z}_{\texttt{summary}}^{\texttt{coop}}$:\\
        \one{This is a very nice \underline{sandal}} that is \two{comfortable and \underline{supportive}}. The only problem is that the straps are \three{a little tight in the toe area}, but it's not a problem. They are \two{very comfortable} and look great with a pair of shoes and dress shoes. \four{Love them}!\\
        \end{mdframed}
    \caption{Example of summaries generated by \name{} with \simpleavg{} and \aggname{} for reviews about a product on the \amazon{} dataset.
    The colors denote the corresponding opinions, and struck-through reviews in gray were not selected by \aggname{} for summary generation (Note that \simpleavg{} uses all the input reviews.) Terms that are more specific to the entity are underlined. Red and struck-through text denotes hallucinated content that has the opposite meaning compared to the input.}
    \label{fig:example-amazon}
\end{figure*}

\begin{table*}[ht]
    \centering
    \small
    \begin{tabular}{cc}
        \toprule
       \textbf{Computing infrastructure} & TITAN V\\
       \midrule
       \textbf{Training duration} & Yelp: 15 hours, Amazon: 12 hours\\
       \midrule
       \textbf{Search strategy} & Manual tuning \\\midrule
       \textbf{Model implementation} & \url{https://github.com/megagonlabs/coop}\\
       \bottomrule
    \end{tabular}

    \vspace{3mm}\begin{tabular}{ccc}
    \toprule
    \textbf{Hyperparameter} & \textbf{Search space} & \textbf{Best assignment} \\
    \midrule
    number of training steps & 100,000 & 100,000\\
    \midrule
    batch size & 256 & 256\\
    \midrule
    tokenizer & \texttt{sentencepiece} & \texttt{sentencepiece} \\
    \midrule
    vocabulary size & 32000 & 32000\\
    \midrule
    embedding size & 512 & 512\\
    \midrule
    encoder & BiLSTM & BiLSTM \\
    \midrule
    hidden size of encoder & 256 & 256 \\
    \midrule
    pooling & \emph{choice}[last, max, mean, self-attn] & mean\\
    \midrule
    number of layers & \emph{choice}[1, 2] & 1 \\
    \midrule
    prior distribution & $\mathcal{N}(\boldsymbol{0}, \boldsymbol{I})$ & $\mathcal{N}(\boldsymbol{0}, \boldsymbol{I})$ \\
    \midrule
    size of latent code & 512 & 512 \\
    \midrule
    decoder & LSTM & LSTM\\
    \midrule
    hidden size of decoder & 512 & 512 \\
    \midrule
    free bits & \emph{choice}[0.0, 0.1, 0.25, 0.5, 1.0, 2.0] & 0.25\\
    \midrule
    KL annealing strategy & \emph{choice}[Cyclic, Pretrain+Anneal] & Pretrain+Anneal\\
    \midrule
    learning rate scheduler & linear schedule with warmup & linear schedule with warmup\\
    \midrule
    learning rate optimizer & Adam & Adam\\
    \midrule
    Adam $\beta_1$ & 0.5 & 0.5\\
    \midrule
    Adam $\beta_2$ & 0.999 & 0.999\\
    \midrule
    learning rate & \emph{choice}[1e-5, 1e-4, 1e-3] & 1e-3 \\
    \midrule
    gradient clip & 5.0 & 5.0 \\
    \bottomrule
    \end{tabular}
    \caption{\name{} search space and the best assignments on \yelp{} and \amazon{} datasets.}
    \label{tab:bimeanvae_param}
\end{table*}

\begin{table*}[ht]
    \centering
    \small
    \begin{tabular}{cc}
        \toprule
       \textbf{Computing infrastructure} & TITAN V\\
       \midrule
       \textbf{Training duration} & Yelp: 25 hours, Amazon: 20 hours\\
       \midrule
       \textbf{Search strategy} & Manual tuning \\\midrule
       \textbf{Model implementation} & \url{https://github.com/megagonlabs/coop}\\
       \bottomrule
    \end{tabular}

    \vspace{3mm}\begin{tabular}{ccc}
    \toprule
    \textbf{Hyperparameter} & \textbf{Search space} & \textbf{Best assignment} \\
    \midrule
    number of training steps & 500,000 & 500,000\\
    \midrule
    batch size & 4 & 4\\
    \midrule
    encoder & \texttt{bert-base-cased} & \texttt{bert-base-cased} \\
    \midrule
    decoder & \texttt{gpt2} & \texttt{gpt2}\\
    \midrule
    prior distribution & $\mathcal{N}(\boldsymbol{0}, \boldsymbol{I})$ & $\mathcal{N}(\boldsymbol{0}, \boldsymbol{I})$ \\
    \midrule
    size of latent code & 512 & 512 \\
    \midrule
    free bits & \emph{choice}[0.0, 0.1, 0.25, 0.5, 1.0, 2.0] & 2.0\\
    \midrule
    KL annealing strategy & \emph{choice}[Cyclic, Pretrain+Anneal] & Cyclic\\
    \midrule
    learning rate scheduler & linear schedule with warmup & linear schedule with warmup\\
    \midrule
    learning rate optimizer & Adam & Adam\\
    \midrule
    Adam $\beta_1$ & 0.5 & 0.5\\
    \midrule
    Adam $\beta_2$ & 0.999 & 0.999\\
    \midrule
    learning rate & \emph{choice}[1e-5, 1e-4, 1e-3] & 1e-5 \\
    \midrule
    gradient norm & 1.0 & 1.0 \\
    \bottomrule
    \end{tabular}
    \caption{Optimus search space and the best assignments on \yelp{} and \amazon{} datasets.}
    \label{tab:optimus_param}
\end{table*}

\end{document}